\documentclass{article} % For LaTeX2e
\usepackage{iclr2027_conference,times}

\usepackage{amsmath,amsfonts,bm}

\def\eqref#1{equation~\ref{#1}}
\def\1{\bm{1}}

\DeclareMathAlphabet{\mathsfit}{\encodingdefault}{\sfdefault}{m}{sl}
\SetMathAlphabet{\mathsfit}{bold}{\encodingdefault}{\sfdefault}{bx}{n}

\usepackage{hyperref}
\usepackage{url}
\usepackage{soul}
\usepackage[most]{tcolorbox}   % ← 这行,要在 begin document 上面
\usepackage[framemethod=tikz]{mdframed}
\usepackage{booktabs}
\usepackage{tabularx}
\usepackage{multirow}
\usepackage{array}
\usepackage[table]{xcolor}
\usepackage{xspace}
\usepackage{graphicx}
\usepackage{lipsum} % Load the package
\usepackage{subcaption}  
\usepackage{caption}   
\usepackage{makecell}
\usepackage{enumitem}
\usepackage{wrapfig}

\setuldepth{foobar}

\title{SD-Cache: Learnable Memory Consolidation with State-Guided Dynamic Caching}

\author{\fontsize{9pt}{10.8pt}\selectfont
Yuanhong Chen$^{1}$,
Yiping Ji$^{1}$,
Cameron Gordon$^{1}$,
Emily Hackett-Jones$^{2}$,
\\
\fontsize{9pt}{10.8pt}\selectfont \bf
Hemanth Saratchandran$^{1}$,
Anton van den Hengel$^{1}$ 
\\[0.5em]
$^{1}$University of Adelaide \\
$^{2}$Commonwealth Bank of Australia
}

\iclrfinalcopy % Uncomment for camera-ready version, but NOT for submission.
\begin{document}

\maketitle

% \begin{abstract}
% The abstract paragraph should be indented 1/2~inch (3~picas) on both left and
% right-hand margins. Use 10~point type, with a vertical spacing of 11~points.
% The word \textsc{Abstract} must be centered, in small caps, and in point size 12. Two
% line spaces precede the abstract. The abstract must be limited to one
% paragraph.
% \end{abstract}

\begin{abstract}
Large language models (LLMs) are increasingly deployed as persistent services that must reason over large volumes of external knowledge. Retrieval-augmented generation (RAG) grounds model responses in relevant documents, but the retrieved content must still be encoded and stored in the key-value (KV) cache. As this content accumulates, the cache grows linearly, increasing memory consumption and inference cost. Existing memory-compression methods address this problem either by retaining a subset of cached states or by learning context compressors that map documents to compact latent representations. Despite these advances, consolidating information under tight memory budgets while preserving its usefulness for future queries remains challenging. We investigate how the model's native KV states can be selectively consolidated into a fixed-size working memory without access to future queries. To this end, we introduce SD-Cache (\textbf{S}tate-guided and \textbf{D}ynamically updated Cache), a query-agnostic KV-memory consolidation framework that operates alongside a frozen generator LLM. SD-Cache learns to consolidate layer-wise KV states into fixed-size memory through selective anchor-based pooling and state-guided refinement, processing new contexts in a single forward pass without context-specific optimisation. Experiments on retrieval-augmented question answering and synthetic long-context benchmarks show that SD-Cache maintains competitive downstream performance under constrained memory budgets.
\end{abstract}
% Existing memory-compression methods address this problem either by retaining only a subset of cached states or by learning context compressors that map documents to compact latent representations. 
% However, these methods often rely on heuristic strategies or treat compressed memory as a static representation, limiting their ability to adapt as new information arrives. 
% % 
% We ask whether a model can instead learn to maintain a budget-constrained working memory that selectively consolidates incoming information and preserves a compact representation of its evolving computational state, without access to future queries. 
% 

% Despite these advances, preserving information useful for future queries under tight memory budgets remains challenging.
% 
% We explore a complementary perspective: learning to selectively consolidate the model’s native KV states into a compact working memory that preserves information useful for subsequent generation, without access to future queries during compression.
% 
% \vspace{-10pt}

\section{Introduction}
\label{sec:intro}

%%%
% Long-running LLM systems → evolving external knowledge → RAG → growing and repeatedly encoded KV cache → existing compression methods are largely static → **need fixed-budget online memory consolidation** → SD-Cache
%%%

% Starter
% Large Language Models (LLMs) have demonstrated remarkable capabilities as general-purpose systems, powering a wide range of applications, including virtual customer-service assistants~\citep{chkirbene2024large,hong2025augmenting}, code generation~\citep{yang2025qwen3,chen2021evaluating}, and mathematical problem solving~\citep{hendrycks2021measuring,lewkowycz2022solving}. 
% Many of these models are now deployed as interactive services that must draw on large and evolving bodies of domain-specific knowledge over extended interactions. 
% To provide access to this knowledge, such systems commonly rely on retrieval-augmented generation (RAG)~\citep{lewis2020retrieval}, which retrieves relevant information from external knowledge bases and incorporates it into the model context. The retrieved content is then encoded into key–value representations that are retained in the KV cache.
Large language models (LLMs) have demonstrated remarkable capabilities as general-purpose systems, powering a wide range of applications, including virtual customer-service assistants~\citep{chkirbene2024large},
code generation~\citep{yang2025qwen3}, and mathematical problem solving~\citep{hendrycks2021measuring,lewkowycz2022solving}.
Many are deployed as interactive services that rely on retrieval-augmented generation (RAG)~\citep{lewis2020retrieval} to access evolving domain-specific knowledge, with retrieved content encoded and retained in the key-value (KV) cache.

% Problems
% This reliance on KV-cache storage creates a fundamental scaling bottleneck. As the cache grows linearly with context length, adding more retrieved content to the model context increases memory consumption and reduces inference throughput. At 16-bit precision, for example, the KV cache required to process a single 128k-token context with LLaMA 70B occupies 84 GB. Similarly, increasing the context length of LLaMA 8B from 1k to 120k tokens reduces its peak throughput on a single H100 GPU by a factor of 77~\citep{grattafiori2024llama,eyuboglu2026cartridges}. 
% These costs are compounded in real-world deployments, where RAG-based customer-service systems may retrieve and re-encode overlapping policy documents and database records across successive interactions~\citep{labruna2025task,tian2022q}, increasing pre-fill latency and introducing redundant information into the KV cache. 
% Hence, Compressing the model context offers a practical way to limit memory and computational costs as information accumulates over time.

This reliance on KV-cache storage creates a scaling bottleneck: the cache grows linearly with context length, increasing memory consumption and limiting inference throughput~\citep{grattafiori2024llama,eyuboglu2026cartridges}. 
% For example, increasing the context length of LLaMA 8B from 1k to 120k tokens reduces peak throughput on a single H100 GPU~\citep{grattafiori2024llama,eyuboglu2026cartridges}. 
These costs are compounded across successive interactions, where customer-service systems may repeatedly retrieve and re-encode overlapping policy documents and database records, increasing pre-fill latency~\citep{labruna2025task,tian2022q}. 
Context compression therefore offers a practical way to reduce the KV storage required for subsequent generation.

\begin{figure}[t]
    \centering
    \includegraphics[width=0.98\linewidth]{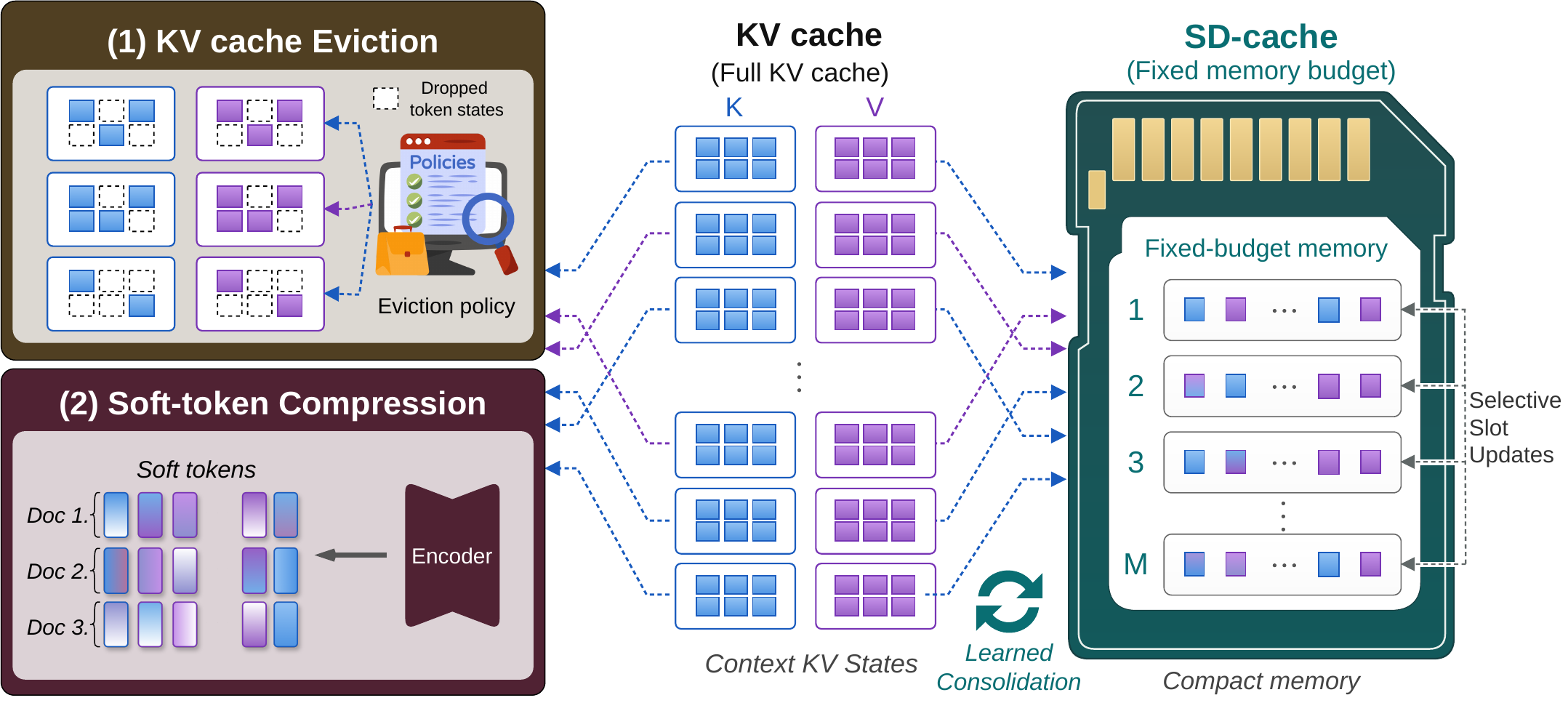}
    \vspace{-8pt}
    \caption{Conceptual comparison of SD-Cache with existing training-free (e.g., KV cache eviction~\citep{li2024snapkv}) and learning-based compression (e.g., soft-token compression~\citep{louis2025pisco}) paradigms.
    SD-Cache instead learns to consolidate KV states into a fixed-budget memory, retaining model-native representations while supporting selective layer-wise updates.
    }
    \vspace{-15pt}
    \label{fig:intro_fig}
\end{figure}

% Hemanth rewrite, use this as your intro is way too long.
Existing KV-cache compression methods fall broadly into training-free and learning-based approaches. 
Training-free methods commonly retain a subset of KV states using positional heuristics~\citep{xiao2024efficient} or attention-based importance estimates~\citep{zhang2023h2o,li2024snapkv}. 
Beyond token selection, context-specific optimisation constructs compact KV representations that approximate the attention behaviour of the original cache~\citep{zweiger2026fast,eyuboglu2026cartridges}.
Although these methods require no additional model training, they rely on predefined heuristics or context-specific optimisation. Under aggressive compression (e.g., above $\times16$), token-selection methods may suffer substantial information loss, as information carried exclusively by discarded states is lost rather than consolidated into the retained cache~\citep{chari2025kv}. 

Learning-based approaches aim to amortise the cost of context processing across multiple queries through reusable compressed representations.
These representations are either optimised for a specific corpus~\citep{eyuboglu2026cartridges} or generated by shared compressors that can be applied to unseen documents \citep{rau2024context,chevalier2023adapting,o2026still}.
Such compressors can encode documents as continuous embeddings~\citep{louis2025pisco} or generate document-specific parameter updates through hypernetworks~\citep{charakorn2026doc,ha2016hyper,hu2021lora}.
% 
% However, these methods focus on constructing compressed representations of individual contexts, which are subsequently reused as static memory units. 
% Incorporating additional information may involve regenerating or re-optimising the representation, or appending new representations, with less emphasis on 
% selectively consolidating information into an existing memory under a fixed budget.
However, approaches based on soft tokens or document-specific parameter updates learn alternative representations of the input context for subsequent reuse, rather than directly
consolidating the layer-wise KV states already formed by the frozen LLM. 
% These native states provide a basis for selectively consolidating contextual information into a fixed-size memory.

In neuroscience, memory consolidation refers to the time-dependent stabilisation and reorganisation of newly encoded memories into durable representations~\citep{squire2015memory}. 
Inspired by this process, we explore context compression for RAG as a memory-consolidation problem.
Specifically, we ask: \textbf{\textit{Can we learn to maintain a fixed-size KV working memory by selectively consolidating contextual information, without access to future queries?}}
To address this question, we introduce SD-Cache, a query-agnostic framework that augments a frozen generator LLM with an auxiliary Transformer-based network (Fig.~\ref{fig:intro_fig}). 
% SD-Cache consolidates the layer-wise KV states of the input context into a fixed set of $M$ memory slots in a single write.
SD-Cache learns a consolidation operation that dynamically constructs compact KV memory for new contexts in a single forward pass, without context-specific optimisation.
Motivated by the clustering structure observed in per-head KV representations, 
% we design an anchor-pooling module (APM) that initialises $M$ memory slots from the KV states of the first $M$ tokens and uses their evolving states to query incoming chunks, 
we design an anchor-pooling module (APM) that initialises $M$ memory slots from the KV states of the first $M$ tokens and refines them through novelty-guided pooling and gated updates,
progressively consolidating the sequence within a fixed memory budget. The resulting attention weights jointly pool the incoming keys and values, preserving key-value correspondence as the slots evolve. 
% 
% As this local pooling may overlook associations spanning the broader context,
% a recurrent state-guided correction module accumulates key-value information directly from the original KV state and uses it to refine the updated slots. SD-Cache thereby maintains a query-agnostic memory whose size remains fixed as the context grows.
To capture key–value associations that slot-wise pooling may overlook, 
a recurrent state-guided correction module builds an associative state directly from the original KV states and uses it to refine the updated slots. SD-Cache thereby maintains a query-agnostic memory whose size remains fixed as the context grows.
Our main contributions are:
\begin{itemize}%[nosep]
% \vspace{-8pt}
\item We formulate query-agnostic KV-cache compression as a learnable memory-consolidation problem and introduce SD-Cache, which maintains a fixed memory budget while leaving the generator LLM frozen.
% \item We develop a two-stage consolidation mechanism that combines anchor-based pooling, which selectively integrates incoming key-value pairs, with recurrent state-guided correction, which captures chunk-level key-value associations that slot-wise pooling may miss.
\item We develop a two-stage consolidation mechanism that combines anchor-based pooling to selectively integrate key-value pairs with state-guided correction to capture chunk-level key-value associations that slot-wise pooling may overlook.
\item We demonstrate on RAG-based question answering~\citep{rau2024bergen,louis2025pisco} and synthetic long-context benchmarks~\citep{hsieh2024ruler,yang2026indexmem} that SD-Cache maintains strong downstream performance at high compression ratios.
\end{itemize}

\section{Related Work}
% \hemanth{This related work is very thorough and in general gives the reader a good exposition of other works that are useful in understanding your paper. However, I do feel it is a bit too detailed and long for a conference paper. It might be worth thinking about how you can summarise some of it so it is not so long. Currently it's about a page and a half which is quite a lot for a related work section.}
% \yuanhong{I have shortened the text and revised the presentation of the descriptions. The text can be further reduce if necessary.}
% Transformer-based language models~\citep{vaswani2017attention,yang2025qwen3,chkirbene2024large} process contextual information through stacked self-attention layers, in which each token is projected into query, key, and value representations~\citep{vaswani2017attention}. During auto-regressive inference, previously computed key and value states are stored in a KV cache, allowing each new token to attend to the preceding context without recomputing earlier activations~\citep{ge2024model} without duplicated computation. As the context length grow with the computational task, the redundancy of the KV cache became a concern in multiple application scope~\citep{louis2025pisco}. Maintain a computational efficient working memory has raised its attention in the field.
% 
\textbf{KV Cache Eviction Methods.} KV cache eviction methods~\citep{xiao2024efficient,zhang2023h2o,oren2024transformers,li2024snapkv,cai2024pyramidkv,yang2026indexmem,gould2026back} formulate memory management as an active forgetting~\citep{davis2017biology} problem, typically implemented through a scoring-and-selection rule. Given cached keys and values $\mathbf{K},\mathbf{V}\in\mathbb{R}^{L\times H\times D}$, where $L$, $H$, and $D$ denote the sequence length, number of KV heads, and per-head feature dimension, respectively, an eviction policy assigns each cache position $i$ an importance score $s_i\in\mathbb{R}$. Under a budget of $M<L$ positions, the retained indices are selected as $\mathcal{I}_M=\operatorname{TopM}(\{s_i\}_{i=1}^{L})$, yielding $\widetilde{\mathbf{K}}=\mathbf{K}[\mathcal{I}_M]$ and $\widetilde{\mathbf{V}}=\mathbf{V}[\mathcal{I}_M]$. 
Retention policies typically use positional~\citep{xiao2024efficient}, attention-based~\citep{zhang2023h2o,oren2024transformers,li2024snapkv}, or predictability-based signals~\citep{gould2026back}. 
Beyond retention criteria, cache budgets can be allocated unevenly across layers to accommodate differences in attention sparsity~\citep{cai2024pyramidkv}, while information from evicted tokens can be encoded into an auxiliary latent memory~\citep{yang2026indexmem}. 
These strategies preserve selected token states or compensate for eviction losses, but do not learn to consolidate information from evicted states directly into the retained KV cache.

\textbf{Learnable Context Compression Methods.}
Learnable compression maps a context $X=(x_1,\ldots,x_L)$ to a compact representation $\mathbf{Z}=f_{\theta}(X)=(\mathbf{z}_1,\ldots,\mathbf{z}_M)$, where $M\ll L$ and $f{\theta}$ is a trainable compressor.
Many methods encode contexts into soft embeddings or memory tokens that condition subsequent generation~\citep{chevalier2023adapting,ge2023context,cheng2024xrag,rau2024context,louis2025pisco}.
Others operate directly on internal KV representations~\citep{eyuboglu2026cartridges}, including approaches that combine learned token selection with parameter-efficient adaptation to consolidate information into retained states~\citep{chari2025kv}.
Concurrent work uses per-layer compactors to synthesise compact KV states from the original cache~\citep{o2026still}. 
While these approaches achieve context compression, our work focuses on learning to selectively consolidate incoming KV states into an existing memory under a fixed budget with a single pass.

\section{Method} \label{sec:method}

\begin{figure*}[t]
    \centering
    \includegraphics[width=0.98\linewidth]{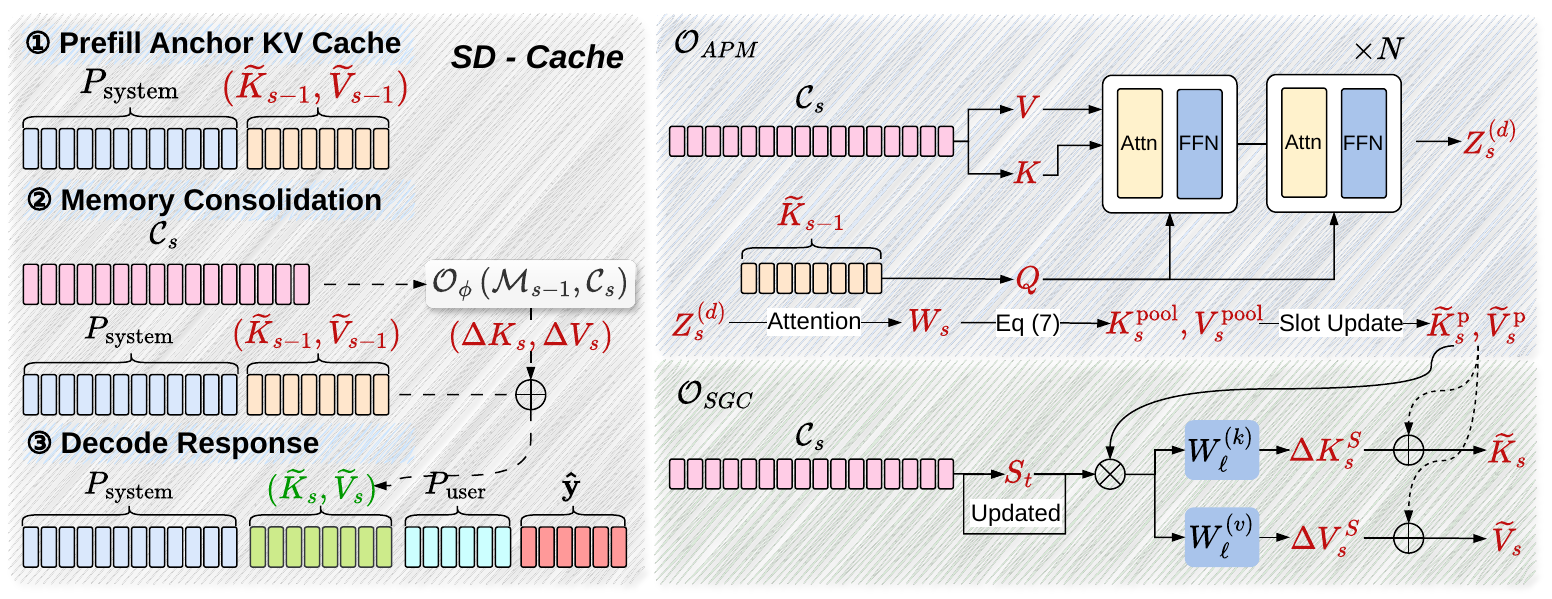}
    \vspace{-8pt}
    \caption{
        Illustration of our SD-Cache method.
        The overall SD-Cache pipeline can be summarised in three major steps when starting from scratch:
        1) initialise the memory with the KV states of the first $M$ document tokens, which serve as anchors;
        2) selectively update the existing memory $\mathcal{M}_{s-1}$ as new context KV states $\mathcal{C}_s$ arrive, while maintaining a fixed budget of $M$ slots; and
        3) use the updated memory $\mathcal{M}_s$ together with the system and user prompts to generate a response with the frozen decoder.
        The right panel details the memory update mechanism: $\mathcal{O}_{APM}$ performs novelty-guided pooling and gated slot updates, while $\mathcal{O}_{SGC}$ refines the pooled KV representations using an associative state accumulated from the original KV stream.
        % The right panel illustrates how $\mathcal{O}_{\mathrm{APM}}$ pools incoming KV states and $\mathcal{O}_{\mathrm{SGC}}$ refines the resulting memory using accumulated key-value associations.
    }
    \label{fig:method}
    \vspace{-15pt}
\end{figure*}

Given a frozen $L$-layer Transformer, a context of length $T$ induces per-layer key and value states
$K^{(\ell)}, V^{(\ell)} \in \mathbb{R}^{H_{\mathrm{kv}} \times T \times d_h}$
where $\ell = 1,\ldots,L$, $H_{\mathrm{kv}}$ denotes the number of KV heads and $d_h$ the head dimension. 
We denote the full KV state induced by the context as
$\mathcal{C} = \left\{\left(K^{(\ell)},V^{(\ell)}\right)\right\}_{\ell=1}^{L}.$
A \emph{memory consolidation operator} transforms the growing KV state into a compact memory
$\mathcal{M}=
\left\{\left(\widetilde K^{(\ell)},\widetilde V^{(\ell)}\right)\right\}_{\ell=1}^{L},$
where
$\widetilde K^{(\ell)},\widetilde V^{(\ell)}\in\mathbb{R}^{H_{\mathrm{kv}} \times M \times d_h}$ and $M \ll T$.
Formally, we define a consolidation operator
\begin{equation}
    \mathcal{O}_{\phi}:\mathcal{C}\rightarrow\mathcal{M}, 
    \qquad 
    \mathcal{M} = \mathcal{O}_{\phi}(\mathcal{C}),
\end{equation}
which maintains a fixed memory budget while preserving the information required for future model computation. 
% Given an unknown future query $Q$, the desired consolidated memory should induce model behaviour close to that obtained from the full KV state:
% \begin{equation}
%     \phi^{*} = \arg\min_{\phi} \;
%     \mathbb{E}_{Q\sim\mathcal{D}_{Q}}
%     \left[
%     D\left(p_{\theta}(\cdot\mid Q,\mathcal{C}),p_{\theta}(\cdot\mid Q,\mathcal{O}_{\phi}(\mathcal{C}))\right)
%     \right],
% \end{equation}
Given an unknown future query $Q$, let
$Y_{\mathcal{C}}=(y_1,\ldots,y_N)=
\operatorname{Decode}\left(p_{\theta}(\cdot\mid \mathcal{C},Q)
\right)$ 
denote the output produced by the frozen Transformer using the full KV state.
The desired consolidated memory should induce model behaviour close to that obtained from the full KV state:
\begin{equation}
    \phi^{*} = \arg\min_{\phi}\;\mathbb{E}_{Q\sim\mathcal{D}_{Q}}
    \left[
    -\sum_{i=1}^{N} \log p_{\theta} \left(y_i \mid y_{<i},\mathcal{O}_{\phi}(\mathcal{C}),Q
    \right)
    \right],
\end{equation}
where $\mathcal{D}_{Q}$ denotes the distribution of future queries.
% where $D(\cdot,\cdot)$ measures the discrepancy between the predictive behaviours induced by the full and consolidated KV states.
% 
Importantly, consolidation is designed to be \emph{query-agnostic}: the future query $Q$ is not available to $\mathcal{O}_{\phi}$ when $\mathcal{M}$ is constructed, as depicted in Fig.~\ref{fig:method}.
% 
% Unlike KV cache eviction methods~\citep{oren2024transformers,li2024snapkv,cai2024pyramidkv,yang2026indexmem,gould2026back,zweiger2026fast} 
% and 
% Building on the one-shot~\citep{chevalier2023adapting,ge2023context,cheng2024xrag,louis2025pisco} formulation above, we extend memory consolidation to update an existing memory with incoming KV states while keeping the memory budget fixed, as depicted in Fig.~\ref{fig:method}.
%
Let $\mathcal{C}_{s}$ denote the KV states induced by the incoming context chunk at update step $s$, and let $\mathcal{M}_{s-1}$ denote the consolidated memory retained from the preceding context. For the arriving context, we extend this formulation by allowing the operator to update the existing consolidated memory with newly observed KV states
\begin{equation}
\mathcal{M}_{s} = \mathcal{O}_{\phi} \left( \mathcal{M}_{s-1}, \mathcal{C}_{s} \right),
\end{equation}

where $\mathcal{O}{\phi} = \mathcal{O}_{\mathrm{SGC}} \circ \mathcal{O}_{\mathrm{APM}} $ and $\mathcal{M}{s}$ remains constrained to a fixed budget of $M$ slots for all $s$.
% A write denotes one application of the consolidation operator. Our main experiments use a single write, with “dynamic” referring to content-dependent memory updates.

% \textbf{Anchor-Pooling Module (APM)}
% The APM method is motivated by an empirical observation on the geometry of Transformer KV states. As illustrated in Fig.~\ref{fig:apm_motivation}, key representations form highly structured, head-specific clusters that remain readily identifiable across layers. Similar structure is also observed for query states. These observation suggest that the model's native KV states provide a more natural basis for memory construction than arbitrary latent slots.

\subsection{Anchor-Pooling Module (\texorpdfstring{$\mathcal{O}_{\mathrm{APM}}$}{O-APM})}
\begin{figure}[t]
    \centering
    \begin{subfigure}[t]{0.38\textwidth}
        \centering
        \includegraphics[width=\linewidth]{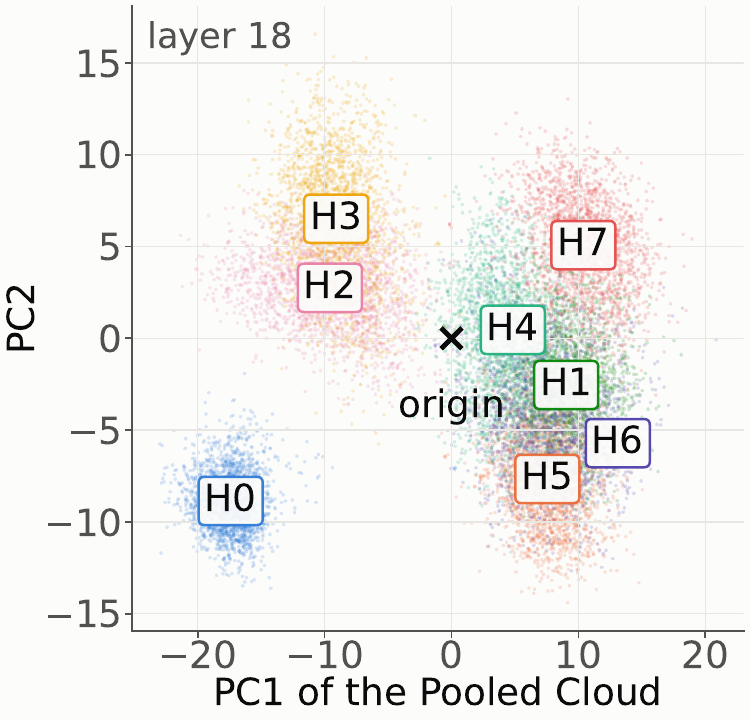}
        \caption{PCA for the Pre-RoPE Keys feature.}
        \label{fig:head_cluster_pca}
    \end{subfigure}\hfill
    \begin{subfigure}[t]{0.6\textwidth}
        \centering
        \includegraphics[width=\linewidth]{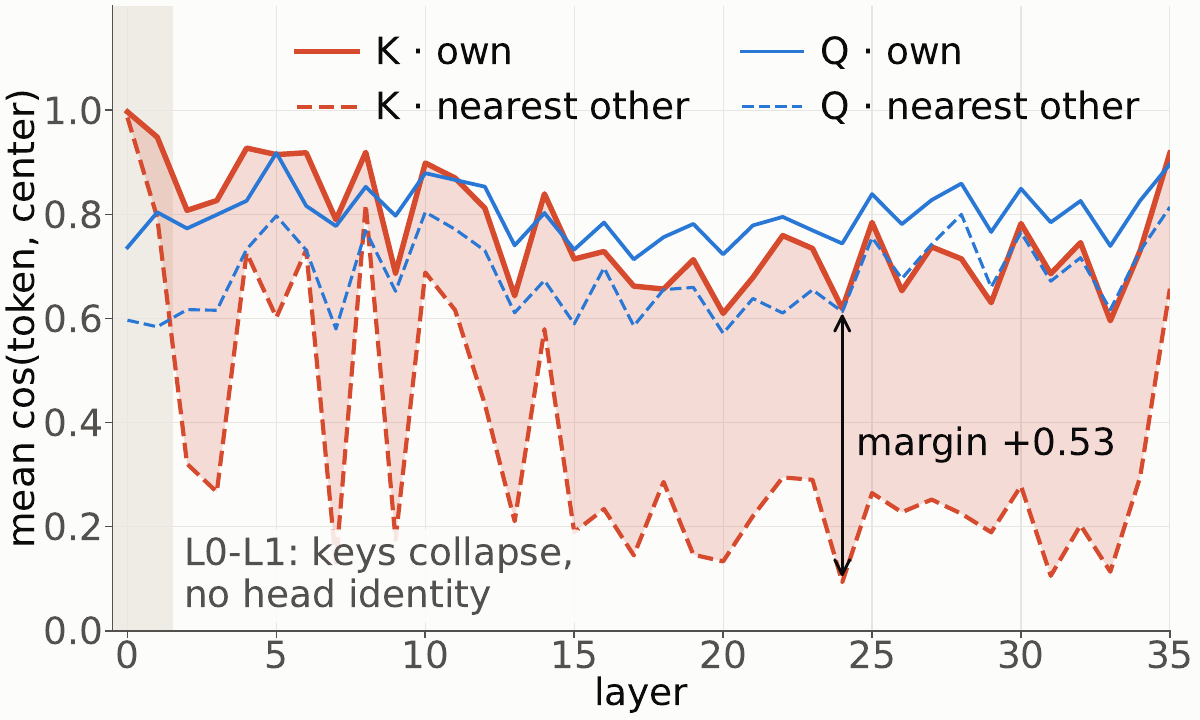}
        \caption{Similarity to Own vs.\ Nearest Other Head Centre.}
        \label{fig:head_margin}
    \end{subfigure}
    \vspace{-5pt}
    \caption{
    % The KV geometry that motivates the $\mathcal{O}_{APM}$. ~\ref{fig:head_margin}
    Visualisation of the PCA projection~\citep{abdi2010principal} of pre-RoPE keys at layer 18 (Figure~\ref{fig:head_cluster_pca}) and comparison of the mean cosine similarities of keys and queries to their own-head centre and nearest other-head centre across layers (Figure~\ref{fig:head_margin}).
    The plots are produced using Qwen3-4B~\citep{yang2025qwen3} with 9,204 natural-language tokens sampled from GoodWiki~\citep{GoodWiki}. 
    % Together, they show that the representations concentrate around head-specific centres, motivating $\mathcal{O}_{\mathrm{APM}}$.
    }
    \label{fig:apm_motivation}
    \vspace{-15pt}
\end{figure}

% The APM is motivated by an empirical observation on the geometry of Transformer KV states. 
APM pools native KV states into compact memory slots, with its design motivated by the structured geometry of Transformer representations.
As illustrated in Fig.~\ref{fig:apm_motivation}, key representations exhibit highly structured, head-specific geometry that remains readily identifiable across layers, with similar structure observed for query representations. These observations suggest that the model's native KV states provide a natural basis for constructing compact memory representations. 
% 
% We therefore initialise $M$ memory slots directly from the original KV states and treat them as \emph{anchors} that can move within the model's native representation space as additional context is absorbed.
% 
For clarity, we omit the layer and KV-head indices below and describe the update for a single set of KV states. At update step $s$, let $(\widetilde K_{s-1},\widetilde V_{s-1})$ denote the current memory of $M$ slots, and let $(K_s,V_s)$ denote the KV states induced by $\mathcal{C}_s$.
APM follows a \emph{novelty-aware write} procedure consisting of novelty estimation, anchor pooling, and selective slot update.
% 
% We assume that at the beginning of the compression we have $s\leftarrow1$, where $s-1$ represent the initial state of $\mathcal{M}$. For the empirical study, we simply use the first $M$ token's KV cache state as the initialisation of the $\mathcal{M}$.
% 
We index the first update by $s=1$ and initialise $\mathcal{M}_0$ using the KV states of the first $M$ context tokens, which serve as the initial anchors.
% We index the first update by $s=1$ and initialise $\mathcal{M}_0$ denotes the initial memory state. In our experiments, we initialise $\mathcal{M}_0$ using the KV states of the first $M$ context tokens, which serve as the initial anchors.

\textbf{Coverage-Aware Novelty Estimation.}
% Before modifying the memory, each incoming key first queries the existing anchors to estimate how well it is already represented. For an incoming key $K_s$, we compute $q_t=\mathrm{LN}(k_t)$ and reconstruct it from a stop-gradient copy of the current memory:
Before modifying the memory, each incoming key first queries the existing memory to estimate how well it is already represented. For the key $\mathbf{k}_{s}$ in the incoming key states $K_s$, we compute $\mathbf{q}_{s}=\mathrm{LayerNorm}(\mathbf{k}_{s})$
and reconstruct it from a stop-gradient copy of the current memory:
\begin{equation}
    \mathbf{r}_s=\mathrm{Attn}
    \left(
    \mathbf{q}_s, \mathrm{sg}[\widetilde K_{s-1}], \mathrm{sg}[\widetilde K_{s-1}]
    \right).
\end{equation}
A lightweight MLP then predicts a scalar novelty score from the incoming representation, its reconstruction, and their residual:
\begin{equation}
    \mathbf{n}_s = \mathrm{MLP}_{\mathrm{nov}}
    \left(
    [\mathbf{q}_s;\,\mathbf{r}_s;\,\mathbf{q}_s-\mathbf{r}_s]
    \right).
\end{equation}
Intuitively, an incoming state that is already well represented by the existing anchors should require little additional memory capacity, whereas poorly reconstructed states provide stronger evidence for updating the memory. We stop gradients through the memory during this read operation to prevent the anchors from artificially increasing the reconstruction error to obtain a larger novelty score.

\textbf{Novelty-Guided Pooling.}
% The current memory keys initialise a set of write queries, which are progressively refined by attending to the incoming keys. 
% We then update these queries using a stack of $L_p$ Perceiver-style blocks~\citep{jaegle2021perceiver}. Each block consists of a pre-normalised cross-attention layer followed by a pre-normalised feed-forward network:
The current memory keys initialise a set of write queries, $Z_s^{(0)}=\widetilde K_{s-1}$, which are refined by attending to the incoming keys.
We update these queries using a stack of $L_p$ Perceiver-style blocks~\citep{jaegle2021perceiver}. For the $d$-th block,
\begin{equation}
    Z_s^{(d)} =\mathrm{PerceiverBlock} \left(
    Q=Z_s^{(d-1)}, K=K_s, V=K_s;\, \mathrm{bias}=\lambda n_s
    \right),
\end{equation}
where $\lambda$ is a learnable scalar.
Each block consists of a pre-normalised cross-attention layer followed by a pre-normalised feed-forward network. The novelty scores $\mathbf{n}_s$ are added directly to the cross-attention logits, biasing the write queries towards incoming keys that are poorly represented by the current memory. 
After $L_p$ blocks, $Z_s^{(L_p)}$ forms the refined write queries used to pool the incoming KV states. We compute a single set of attention weights and apply it to both keys and values:
\begin{equation}
    W_s = \mathrm{softmax} \left(
    \frac{Z_s^{(L_p)}K_s^\top}{\sqrt{d_h}} +
    \lambda n_s \right), 
    \, K_s^{\mathrm{pool}}=W_sK_s, \,V_s^{\mathrm{pool}}=W_sV_s.
\end{equation}
Using the same weights for both keys and values ensures that each pooled KV pair aggregates information from the same context positions.

\textbf{Selective Slot Update.}
Finally, a shared per-slot gate determines how much newly pooled information is incorporated into each memory slot:
$\mathbf{g}_s=\sigma\left(
\mathrm{MLP}_{\mathrm{gate}}
\left([\widetilde K_{s-1};\widetilde V_{s-1};K_s^{\mathrm{pool}};V_s^{\mathrm{pool}}]
\right)\right)\in(0,1)^M.$
The same gate is shared between the key and value of each slot, preserving their correspondence:
\begin{equation}
    \widetilde K_s^{\mathrm{p}} = (1-\mathbf g_s)\widetilde K_{s-1} + \mathbf g_s K_s^{\mathrm{pool}},
    \qquad 
    \widetilde V_s^{\mathrm{p}} = (1-\mathbf g_s)\widetilde V_{s-1} + \mathbf g_s V_s^{\mathrm{pool}}.
    \label{eq:apm_slot_update}
\end{equation}
The gate is initialised to favour retaining the existing memory, allowing the slots to gradually learn to incorporate new information during training. The resulting $(\widetilde K_s^{\mathrm{p}},\widetilde V_s^{\mathrm{p}})$ form the provisional consolidated memory produced by APM. 
% While APM organises incoming KV states into a fixed number of representative slots, pooling alone may not fully preserve the global key-value associations present in the original context. We therefore introduce a State-Guided Correction module to complement the pooled memory with information accumulated directly from the original KV states.

\subsection{State-Guided Correction (\texorpdfstring{$\mathcal{O}_{\mathrm{SGC}}$}{O-SGC})}
SGC complements APM with a compact associative state~\citep{yang2024parallelizing,hatamizadeh2026gated} constructed directly from the original, uncompressed KV stream. Importantly, the state is updated from the raw KV states rather than from the pooled memory, providing an information path decoupled from the slot-pooling process. Within each KV head, the state $S_t\in\mathbb{R}^{d_h\times d_h}$ represents a key-to-value associative map accumulated over the original stream.
For each token $t$ in $\mathcal{C}_s$, we first normalise its key as
$\hat{\mathbf{k}}_t=\mathbf{k}_t/\lVert\mathbf{k}_t\rVert_2$
and predict a data-dependent write strength
$\beta_t=\sigma\left(\mathbf{w}_{\beta}^{\top}\mathbf{h}_t+b_{\beta}\right)$,
where $\mathbf{h}_t$ denotes the corresponding hidden representation. 
% 
% Starting from $S_0=0$, we update the associative state using a gated delta rule~\citep{hatamizadeh2026gated}:
Starting from $S_0=0$, the state first absorbs the $M$ anchor tokens with
$\beta=\sigma(b_{\beta})$, and the associative state is then updated using a gated delta rule~\citep{hatamizadeh2026gated}:
\begin{equation}
    S_t = \gamma_t S_{t-1}
    \left(I-\beta_t\hat{\mathbf{k}}_t\hat{\mathbf{k}}_t^{\top}\right)
    +
    \beta_t\mathbf{v}_t\hat{\mathbf{k}}_t^{\top},
    \qquad \gamma_t=1-\beta_t.
\end{equation}
This update continuously absorbs key-value associations from the original KV chunk, while the coupled decay stabilises the state over long sequences.
After processing the KV states, we use the resulting associative state $S$ to refine the updated slots $(\widetilde K_s^{\mathrm{p}},\widetilde V_s^{\mathrm{p}})$ produced by APM. Their keys are first normalised and used to query the state:
\begin{equation}
    R_s^{v} = S(\widehat{\widetilde K}_s)^\top,
    \qquad
    \widehat{\widetilde K}_s = \frac{\widetilde K_s^{\mathrm{p}}}{\lVert\widetilde K_s^{\mathrm{p}}\rVert_2}.
\end{equation}

% Since $S$ represents a key-to-value mapping, $R_s^{v}$ provides the value-side association retrieved by each updated key. 
% Rather than assuming an inverse value-to-key mapping, we transform this single well-defined readout using lightweight learned correction heads:
Since $S$ defines only a key-to-value mapping, we use the same readout $R_s^v$ to predict both key and value corrections through lightweight learned heads:
\begin{equation}
    \Delta V_s^{S} = W_{\ell}^{(v)}R_s^{v},
    \qquad
    \Delta K_s^{S} = W_{\ell}^{(k)}R_s^{v},
\end{equation}
where $W_{\ell}^{(v)}$ and $W_{\ell}^{(k)}$ are zero-initialised layer-specific projections. The corrected memory is then
\begin{equation}
    \widetilde K_s = \widetilde K_s^{\mathrm{p}} + \alpha_k(\Delta K_s^{S})^\top,
    \qquad
    \widetilde V_s = \widetilde V_s^{\mathrm{p}} + \alpha_v(\Delta V_s^{S})^\top.
\end{equation}
The state matrix $S$ is only maintained during the update and is discarded afterwards, leaving no additional persistent state beyond the compact KV memory.

% The zero initialisation allows SD-Cache to start from the APM solution and progressively learn how the associative state should refine the consolidated memory.

% \subsection{Training Objective.}
% We keep the parameters $\theta$ of the underlying Transformer frozen and optimise only the active parameters in $\mathcal{O}_{\mathrm{APM}}$ and $\mathcal{O}_{\mathrm{SGC}}$. 
% The objective of SD-Cache is to replicate the full-cache output answer. Given a training example consisting of context $\mathcal{C}$, consolidated memory $\mathcal{M}$, query $Q$, and target sequence $Y=(y_1,\ldots,y_N)$, the training loss is defined as
% $\mathcal{L}=-\sum_{i=1}^{N}\log p_{\theta}\left(y_i \mid y_{<i},Q,\mathcal{M}\right)$.

\subsection{Training Objective}
We keep the parameters $\theta$ of the underlying Transformer frozen and optimise only the parameters $\phi$ of $\mathcal{O}$. For each training example, we first decode the target sequence $Y_{\mathcal{C}}=(y_1,\ldots,y_N)$ using the full KV state $\mathcal{C}$. SD-Cache is then trained to reproduce this full-cache output using only the consolidated memory $\mathcal{M}$ with $\mathcal{L}=-\sum_{i=1}^{N}\log p_{\theta}\left(y_i \mid y_{<i},Q,\mathcal{M}\right)$.
This objective encourages the estimated memory to preserve the information required to reproduce the behaviour of the full KV.

% \begin{equation}
%     \mathcal{L}=-\sum_{i=1}^{N}\log p_{\theta}
%     \left(
%     y_i \mid y_{<i},Q,\mathcal{M}
%     \right).
% \end{equation}

% Although the ideal objective introduced above is to preserve the predictive behaviour of the full KV state, we do not explicitly distil the full-cache output distribution. 
% Instead, SD-Cache is trained using standard next-token prediction. Given a training example consisting of context $\mathcal{C}$, query $Q$, and target sequence $Y=(y_1,\ldots,y_N)$, we first construct the consolidated memory $\mathcal{M}=\mathcal{O}_{\phi}(\mathcal{C})$ without access to $Q$, and optimise
% \begin{equation}
%     \mathcal{L}=-\sum_{i=1}^{N}\log p_{\theta}
%     \left(
%     y_i \mid y_{<i},Q,\mathcal{M}
%     \right).
% \end{equation}
% This preserves the query-agnostic nature of memory construction while allowing downstream supervision to determine which information must remain recoverable from the compressed KV state.

% \clearpage

\section{Experiments}
\textbf{Evaluation Protocols.} 
Our primary benchmark is retrieval-augmented question answering (RA-QA) Benchmark~\citep{louis2025pisco}. The training data of this benchmark is taken from questions from multiple sources of QA datasets~\citep{louis2025pisco}~\footnote{https://huggingface.co/datasets/maxoul/pisco\_finetuning\_data}, which contain 453,000 samples. Similar to PISCO~\citep{louis2025pisco}, we also use the response from a teacher model~\footnote{https://huggingface.co/mistralai/Mistral-7B-Instruct-v0.2} as the supervision signal of the next token prediction loss. The evaluation includes five popular QA datasets including: Natural Questions~\citep{kwiatkowski2019natural}, TriviaQA~\citep{joshi2017triviaqa}, PopQA~\citep{mallen2023not}, ASQA~\citep{stelmakh2022asqa}, and HotpotQA~\citep{yang2018hotpotqa}. Our main evaluation metric is the answer matching accuracy in the QA context, which we define as correct prediction if the normalised label is found within the normalised prediction~\citep{louis2025pisco}. We adapt Mistral-7B-v0.2~\citep{Jiang2023Mistral7} as the backbone for this benchmark.
We additionally report results on RULER~\citep{hsieh2024ruler} at a context length of $4096$ to further evaluate the model's long-context retrieval and reasoning capabilities, including needle-in-a-haystack (NIAH), multi-hop tracing, and frequency-based aggregation tasks. 
For this benchmark, we adapt and augment the RA-QA training data with synthetic retrieval, frequency, and tracking tasks, and use Qwen3-8B~\citep{yang2025qwen3}, Mistral-7B-v0.3~\citep{Jiang2023Mistral7}, and Llama-3.1-8B~\citep{grattafiori2024llama} as generator backbones.
Following the official protocol we score every task with the reference \texttt{string\_match} implementation and report the unweighted mean over tasks as the aggregated score. Implementation details are provided in Appendix~\ref{aap:imple_details}.

% \textbf{Implementation Details} 
% Our consolidation network follows the Perceiver architecture: the $M$ slot queries cross-attend to the incoming key-value stream through $D=8$ pre-norm blocks, each with $8$ cross-attention heads and a $4\times$ FFN expansion, operating at the backbone's key-value width. 
% One trunk is shared across all layers; each layer additionally owns a lightweight updater and two rank-$16$ LoRA adapted ($\alpha=32$) on the frozen key and value projections. This amounts to approximate $142$M trainable parameters, $1.7\%$ of an 8B backbone, which remains frozen throughout. We set fixed correction strengths $\alpha_v=\alpha_k=0.1$.
% We train for five epochs with AdamW at a learning rate of $1\times10^{-4}$ under a linear schedule with $3\%$ warmup, in bf16, with an effective batch size of $64$ (per-device batch $1$ with gradient accumulation). Training runs under DeepSpeed ZeRO-2 on four RTX 6000 Ada GPUs; the per-layer trunk calls are gradient-checkpointed, which leaves the computed gradients unchanged and is what fits the largest memory budget into $48$\,GB.

\begin{table*}[!tb]
    \centering
    \caption{
        \textbf{Answer-matching accuracy (\%) on five RA-QA datasets~\citep{louis2025pisco}.}
        Using five retrieved documents per query, SD-Cache achieves the highest average accuracy among the compared compression methods at a $16\times$ compression ratio.
        $\dagger$ methods limited to one context. $\ddagger$ methods using Llama-2 7B. $^\star$ use official checkpoint to reproduce the results. 
    }
    \vspace{-10pt}
    \resizebox{0.98\textwidth}{!}{  
    \begin{tabular}{lccccccc}  
        \toprule
        & \textbf{Rate} & \textbf{ASQA} & \textbf{HotpotQA} & \textbf{NQ} & \textbf{TriviaQA} & \textbf{POPQA} & \textbf{Average} \\
        \midrule
        \multicolumn{8}{c}{\textbf{Mistral-7B without Compression}} \\
        \midrule
        Closed-book~\citep{louis2025pisco} & - & 51.00 & 34.00 & 46.00 & 79.00 & 29.00 & 47.80 \\
        Full KV~\citep{louis2025pisco} & - & 74.00 & 51.00 & 69.00 & 92.00 & 70.00 & 71.00 \\
        % Llama-3.1-8B & - & 71.00 & 50.00 & 65.00 & 90.00 & 68.00 & 69.00 \\
        % Solar-10.7B & - & 75.00 & 55.00 & 7100 & 93.00 & 71.00 & 73.00 \\
        \midrule
        \multicolumn{8}{c}{\textbf{Compression Models}} \\
        \midrule
        xRAG$^\dagger$ \citep{cheng2024xrag}	&	x128	&	34.00	&	27.00	&	32.00	&	77.00	&	33.00	&	40.00	\\
        A.C.$^{\dagger\ddagger}$~\citep{chevalier2023adapting}	&	x4	&	57.00	&	31.00	&	35.00	&	70.00	&	24.00	&	43.00	\\
        ICAE$^\ddagger$~\citep{ge2023context}	&	x4	&	47.00	&	29.00	&	42.00	&	78.00	&	43.00	&	48.00	\\
        DODO$^\ddagger$~\citep{qin2024dodo}	&	x10	&	52.00	&	38.00	&	48.00	&	82.00	&	47.00	&	53.00	\\
        COCOM~\citep{rau2024context}	&	x16	&	63.00	&	46.00	&	58.00	&	89.00	&	48.00	&	61.00	\\
        PISCO$^\star$~\citep{louis2025pisco}	&	x16	&	71.09	&	\textbf{48.06}	&	64.58	&	90.17	&	65.83	&	67.94	\\
        % \midrule
        \rowcolor{gray!10}
        SD-Cache (Ours)	&	x16	&	\textbf{73.10}	&	47.48	&	\textbf{65.92}	&	\textbf{90.81}	&	\textbf{66.06}	&	\textbf{68.67}	\\
        \midrule
        PISCO~\citep{louis2025pisco}	&	x128	&	68.00	&	\textbf{46.00}	&	61.00	&	89.00	&	55.00	&	63.80	\\
        % \midrule
        \rowcolor{gray!10}
        SD-Cache (Ours)	&	x128	&	\textbf{71.84}	&	44.75	&	\textbf{63.73}	&	\textbf{89.58}	&	\textbf{57.41}	&	\textbf{65.46}	\\        \bottomrule
        \end{tabular}
        }
    \label{table:raqa_results}
    \vspace{-10pt}
\end{table*}

\begin{figure}[t]
    \centering
    \begin{subfigure}[t]{0.495\textwidth}
        \centering        
        \includegraphics[width=\linewidth]{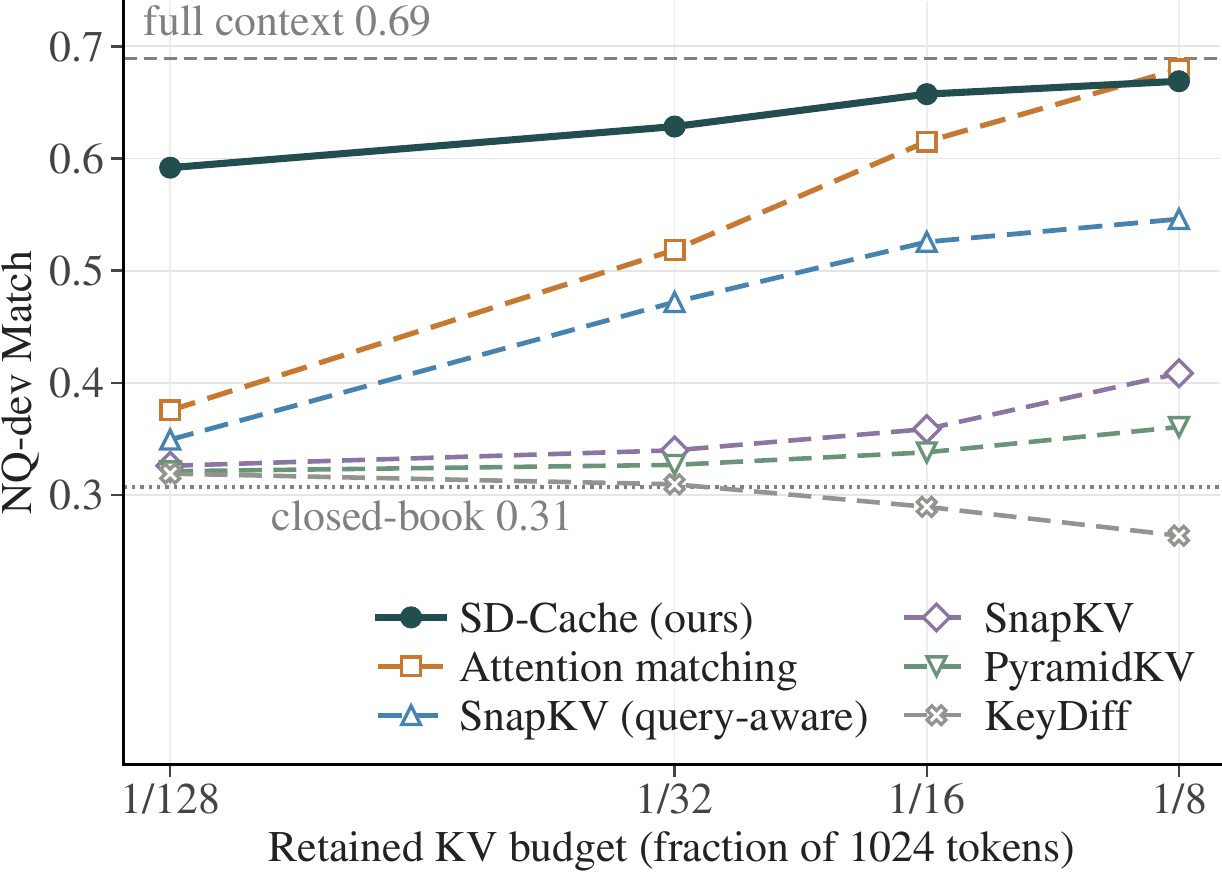}
        \vspace{-10pt}
        \caption{Performance on RA-QA.}
        \vspace{-5pt}
        \label{fig:raqa_tf_match_vs_rate}
    \end{subfigure}
    \begin{subfigure}[t]{0.495\textwidth}
        \centering
        \includegraphics[width=\linewidth]{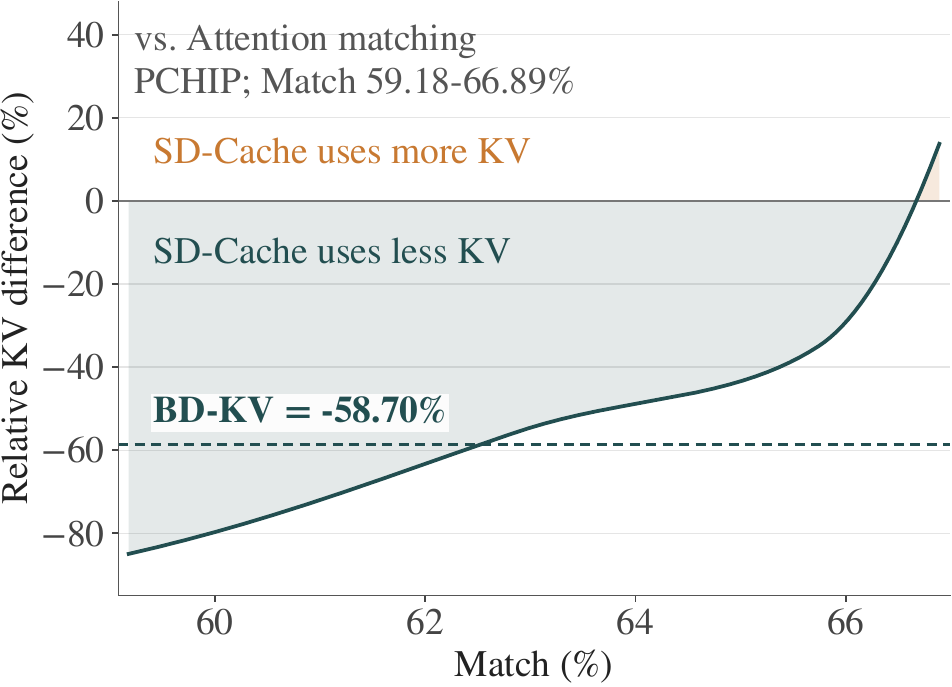}
        \vspace{-10pt}
        \caption{KV Cost at Equal Match Score.}
        \vspace{-5pt}
        \label{fig:nq_bd_kv}
    \end{subfigure}
    \vspace{-5pt}
    \caption{
    Performance comparison on NQ in RA-QA benchmark~\citep{rau2024bergen} under (\ref{fig:raqa_tf_match_vs_rate}) four compression settings (e.g., 
    $1/8$, $1/16$, $1/32$ and $1/128$) and (\ref{fig:nq_bd_kv}) the average relative change in KV budget at matched accuracy. 
    }
    \vspace{-15pt}
    % \label{fig:comparison}
\end{figure}

\subsection{Results on RA-QA Benchmark}
Tab.~\ref{table:raqa_results} compares SD-Cache with existing methods on five RA-QA datasets. At a compression ratio of $16\times$, SD-Cache achieves the highest average accuracy among the compared compression methods, reaching $68.67\%$ and outperforming PISCO by 0.73 percentage points (pp) at the same compression ratio. SD-Cache improves over PISCO on four of the five datasets, with the largest gains on ASQA and Natural Questions ($+2.01$~pp and $+1.34$~pp, respectively), while maintaining comparable performance on other datasets. 
Its average accuracy also remains within approximately $3$~pp of the uncompressed Mistral-7B baseline. 
At $128\times$ compression, SD-Cache also outperforms PISCO~\citep{louis2025pisco} by $1.66$~pp in average accuracy, further highlighting its effectiveness under aggressive compression.
These results indicate that SD-Cache preserves information useful for downstream question answering under substantial context compression.

\textbf{Performance under Different Compression Ratios.}
We further investigate the effect of compression on answer-matching accuracy using Qwen3-4B-Instruct~\citep{yang2025qwen3} on the Natural Questions (NQ)~\citep{rau2024bergen} development set. As shown in Fig.~\ref{fig:raqa_tf_match_vs_rate}, SD-Cache's advantage over Attention Matching~\citep{zweiger2026fast} grows as the compression ratio increases. At $16\times$ and $32\times$ compression, SD-Cache achieves accuracies of $65.74\%$ and $62.8\%$, exceeding Attention Matching by $4.24$~pp and $11.0$~pp. At $128\times$ compression, the gap increases to $21.7$~pp, with SD-Cache retaining $59.2\%$ accuracy compared with $37.5\%$ for Attention Matching. 
% At the lower compression ratio of $8\times$.
% 
We further demonstrate the accuracy-memory trade-off between our method and Attention Matching~\citep{zweiger2026fast} in its fastest configuration. 
We report Bjøntegaard-Delta rate~\citep{bjontegaard2001calculation} applied to KV-cache budgets (BD-KV) to measure the average relative change in KV budget at matched accuracy over the shared quality range, with negative values indicating memory savings.
In Fig.~\ref{fig:nq_bd_kv}, we plot the relative KV budget difference at matched accuracy, where negative values indicate lower KV requirements for SD-Cache. Over the shared Match range of $59.18\%$-$66.89\%$, SD-Cache achieves a BD-KV of $-58.70\%$. Its memory advantage decreases as the target accuracy increases and reverses near the upper end of this range. Details about BD-KV are provided in Appendix~\ref{app:bd-kv}.
These results indicate that SD-Cache is particularly effective under tight memory budgets, but it falls slightly behind Attention Matching at the largest evaluated KV budget.
One possible explanation is that learned compression may lose fine-grained information that could otherwise be retained when more KV capacity is available. 
This motivates exploring an adaptive hybrid method that combines selective KV retention with learned compression under a fixed memory budget in the future.

\begin{table}[t]
    \centering
    \small
    \setlength{\tabcolsep}{4pt}
    \renewcommand{\arraystretch}{1.15}
    \caption{
        Average accuracy (\%) across the 13 RULER-4K tasks~\citep{hsieh2024ruler} under different KV cache reduction ratios, using Llama 3.1-8B~\citep{grattafiori2024llama} as the base model. A reduction ratio of $0\%$ denotes the full, uncompressed KV cache.
    }
    \vspace{-10pt}
    \resizebox{0.97\linewidth}{!}{%
    \begin{tabularx}{\linewidth}{
        l
        >{\raggedright\arraybackslash}X
        rrrrrr
    }
        \toprule
        \multirow{2}{*}{\textbf{Model}}
        & \multirow{2}{*}{\textbf{Method}}
        & \multicolumn{6}{c}{\textbf{KV cache reduction}} \\
        \cmidrule(lr){3-8}
        & & 0.00\% & 75.00\% & 90.00\% & 93.75\% & 96.88\% & 98.44\% \\
        \midrule
        % \multirow{7}{*}{\rotatebox{90}{Llama-3.1-8B~\citep{grattafiori2024llama}}}
        % \multirow{7}{*}{Llama-3.1-8B}
        \multirow{7}{*}{\makecell{Llama\\3.1-8B}}
        & PyramidKV~\citep{cai2024pyramidkv}
        & 95.30 & 43.60 & 23.10 & 16.47 & 8.64 & 0.06 \\
        & SnapKV~\citep{li2024snapkv}
        & 95.30 & 63.20 & 43.40 & 16.45 & 8.60 & 0.06 \\
        & TOVA~\citep{oren2024transformers}
        & 95.30 & 63.30 & 37.50 & 20.28 & 6.83 & 3.70 \\
        & ExpectedAttention~\citep{devoto2025expected}
        & 95.30 & 75.90 & 30.60 & 15.43 & 10.54 & 7.87 \\
        & AM-Fast~\citep{zweiger2026fast} 
        & 95.30	& 78.82	& 46.95	& 33.75	& 22.73	& 12.55 \\
        & KeyDiff~\citep{park2026keydiff}
        & 95.30 & 64.44 & 61.10 & 54.50 & 34.69 & 16.58 \\
        & IndexMem~\citep{yang2026indexmem}
        & 95.30 & 78.30 & 55.20 & -- & -- & -- \\
        % & PISCO~\citep{louis2025pisco}
        % & 95.30 & 63.48 & 55.20 & -- & -- & -- \\
        \cmidrule(l){2-8}
        \rowcolor{gray!10}
        \cellcolor{white}
        & \textbf{SD-Cache (Ours)}
        & 95.30
        & \textbf{79.52}
        & \textbf{66.11}
        & \textbf{55.44}
        & \textbf{55.17}
        & \textbf{53.78} \\
        \bottomrule
    \end{tabularx}
    }
    \vspace{-15pt}
    \label{table:ruler_results}
\end{table}

\textbf{Compaction Efficiency.}
We analytically estimate computational costs in TFLOPs~\citep{louis2025pisco}, separating context preparation (\textit{Compact}, including context pre-fill) from subsequent inference; \textit{Total} sums both stages, shown in Tab.~\ref{tab:main_efficiency}. On Mistral-7B, SD-Cache requires $1.79$ TFLOPs for inference after preparation, while its compaction overhead increases the total cost to $13.95$ TFLOPs, compared with $11.82$ for Full KV and $12.88$ for PISCO. 
\begin{wraptable}{r}{0.45\linewidth}
    % \vspace{-0.5em}
    \centering
    \small
    \setlength{\tabcolsep}{3.5pt}
    \renewcommand{\arraystretch}{1.10}
    \vspace{-2pt}
    \caption{Analytical computational cost.}
    \vspace{-6pt}
    \resizebox{\linewidth}{!}{%
        \begin{tabular}{l*{3}{>{\raggedleft\arraybackslash}p{4.5em}}}
            \toprule
            & \multicolumn{3}{c}{Computational cost (TFLOPs) $\downarrow$} \\
            \cmidrule(lr){2-4}
            Method & Inference & Compact & Total \\
            \midrule

            \multicolumn{4}{l}{\textit{Mistral-7B / RAG-QA}} \\
            FullKV & 1.79 & 10.03 & 11.82 \\
            AutoCompressor & 2.98 & 10.42 & 13.40 \\
            PISCO & 2.89 & 9.99 & 12.88 \\
            \textbf{SD-Cache (Ours)} & 1.79 & 12.16 & 13.95 \\

            \midrule

            \multicolumn{4}{l}{\textit{Llama-3.1-8B / 4K context}} \\
            FullKV & 2.74 & 65.97 & 68.71 \\
            SnapKV & 2.42 & 66.11 & 68.53 \\
            AM-fast & 2.29 & 169.90 & 172.18 \\
            \textbf{SD-Cache (Ours)} & 2.43 & 78.27 & 80.70 \\

            \bottomrule
        \end{tabular}%
    }
    \vspace{-15pt}
    \label{tab:main_efficiency}
    % \vspace{-0.5em}
\end{wraptable}
On Llama-3.1-8B with a 4K context, SD-Cache requires $80.70$ TFLOPs in total, substantially less than AM-fast ($172.18$), but more than SnapKV ($68.53$).
For more results, please refer to Appendix~\ref{app:tflops_detailed}.
These estimates quantify the additional computation required to construct SD-Cache's compact KV representations. When the prepared context can be reused across queries, this preparation cost can be amortised, yielding an average cost of $I + C/N_q$ for $N_q$ queries sharing the same context.

% 
% Fig~\ref{fig:raqa_acc-time} compares answer-matching accuracy against mean compaction time per sample, measured on a single RTX 4090 GPU with a batch size of one. At $16\times$ compression, SD-Cache requires $0.20$ seconds per sample, compared with $2.13$ seconds for Attention Matching, yielding an approximately $10.7\times$ speed-up in compaction while improving accuracy by 3.1 percentage points. PISCO~\citep{louis2025pisco} achieves a lower compaction time of $0.09$ seconds but obtains $60.4\%$ accuracy. SD-Cache improves upon this accuracy by $4.2\%$ at an additional cost of $0.11$ seconds per sample. These results demonstrate a favourable balance between answer quality and compaction cost.
% 
% \input{tabs/tflops_main}
% 

\subsection{Results on RULER Benchmark}

Tab.~\ref{table:ruler_results} reports the average accuracy across the 13 RULER tasks with a 4096 token context using Llama-3.1-8B~\citep{grattafiori2024llama}. SD-Cache achieves the highest reported accuracy at every non-zero cache reduction ratio. 
At $75\%$ and $90\%$ reduction, it surpasses the second-best methods by $0.7$~pp and $5.01$~pp, respectively.
Its advantage is particularly pronounced at the two highest reduction ratios: at $96.88\%$ and $98.44\%$, SD-Cache achieves accuracies of $55.17\%$ and $53.78\%$, exceeding KeyDiff~\citep{park2026keydiff}, the strongest baseline with available results in these settings, by $20.48$~pp and $37.20$~pp, respectively. 
Furthermore, SD-Cache's accuracy decreases by only $1.66$~pp as the cache reduction increases from $93.75\%$ to $98.44\%$. These results indicate that SD-Cache preserves useful information for long-context tasks even under severe KV cache compression. 
See Appendix~\ref{app:ruler_additional} for results on additional backbones.
These results highlight the value of learning to compress KV states while preserving task performance under tight memory budgets. 
Hence, a promising direction for future work is to combine learned consolidation with training-free token selection to draw on the complementary strengths of both approaches.

\subsection{Ablation Studies and Analysis}
\textbf{Contributions of Individual Components.}
We first analyse the contributions of SD-Cache's key components on Natural Questions (NQ) within the RA-QA benchmark, using Qwen3-4B-Instruct-2507\footnote{\url{https://huggingface.co/Qwen/Qwen3-4B-Instruct-2507}} as the backbone at a fixed retention ratio of $1/16$. 
A simple baseline that compresses text into soft tokens achieves a score of $61.72\%$. The anchor pooling module (APM) variant improves this score to $64.44\%$, yielding a gain of $2.72$~pp. Incorporating state-guided correction (SGC) further increases the score to $65.74\%$, providing an additional gain of $1.3$~pp. Full SD-Cache improves upon the soft-token baseline by $4.02$~pp, highlighting the incremental contributions of APM compression and SGC.

\textbf{Length Generalisation under a Fixed Memory Budget.}
We evaluate length generalisation on RULER QA by increasing the context length from 4K to 8K and 16K while keeping the memory budget fixed at $M{=}410$ slots.
Starting from a model trained on 4K-token contexts with a single write, we compare current single write setting with two streaming strategies:
% \emph{recurrent rolling (RR)}, which carries the updated slots forward as the state for the next write, and
\emph{recurrent rolling (RR)}, which passes the updated slots to the next write, and
\emph{skip-connected rolling (SCR)}, which adds skip connections between subsequent writes.
The two strategies perform similarly at 8K with two writes ($26.7\%$ vs. $25.7\%$), but diverge at 16K. 
RR degrades sharply and often produces gibberish responses, with accuracy falling to $9.9\%$ after three updates and $1.7\%$ after four. In contrast, SCR retains $28.5\%$ and $26.9\%$, respectively.
These results show that SD-Cache supports iterative writing, but maintaining accuracy across repeated updates remains challenging, motivating future work on more robust update strategies and training data tailored to multi-step writing.

\begin{figure}[t]
    \centering
    \begin{minipage}[t]{0.45\linewidth}
        \vspace{0pt}
        \centering
        \includegraphics[width=\linewidth]{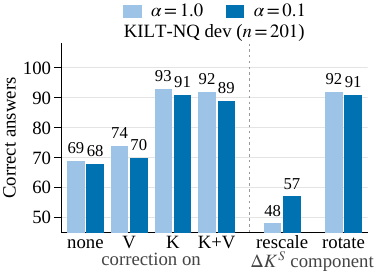}
        % \vspace{-5pt}
        \caption{State-guided correction ablations: correction targets (left) and parallel (``rescale'') versus orthogonal (``rotate'') key-correction components (right).}
        \vspace{-15pt}
        \label{fig:sd_ablation}
    \end{minipage}\quad\quad
    \begin{minipage}[t]{0.45\linewidth}
        \vspace{0pt}
        \centering
        \includegraphics[width=\linewidth]{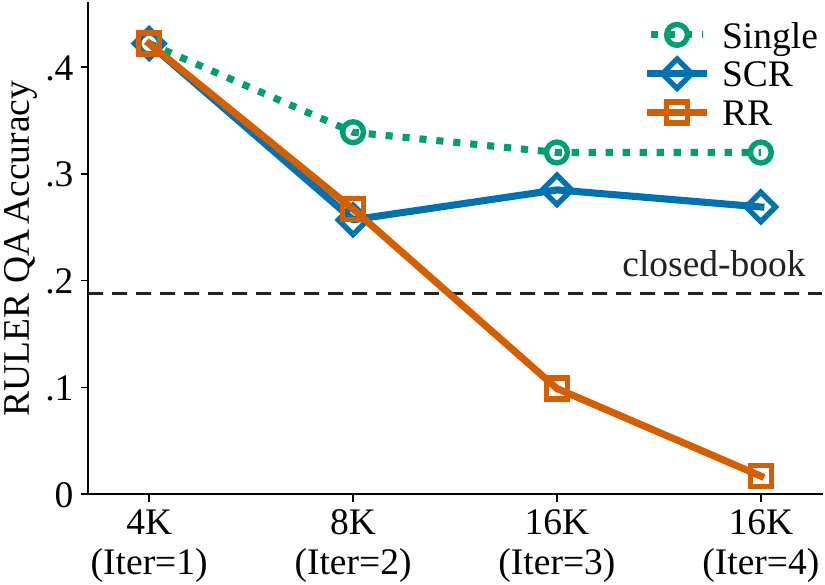}
        % \vspace{-5pt}
        \caption{Length generalisation on RULER QA with $M=410$ slots. RR: recurrent rolling; SCR: skip-connected rolling. ``Iter'' denotes the number of writes.}
        \vspace{-15pt}
        \label{fig:sdcache_vs_ragged}
    \end{minipage}
    % \vspace{-10pt}
\end{figure}

\textbf{Analysis of State-Guided Correction.}
To understand where SGC helps, we ablate guidance strength $\alpha$ and correction target using models trained on RA-QA~\citep{louis2025pisco} and evaluated on 201 NQ samples.
We divide the experiments into two groups (Fig.~\ref{fig:sd_ablation}): 1) applying correction to $K$, $V$, both, or neither; and 2) retaining only the component of the applied key correction parallel (``rescale'') or orthogonal (``rotate'') to the provisional slot key $\tilde{K}$, while keeping value correction unchanged.
The parallel component scales the existing key, whereas the orthogonal component adjusts its direction.
Most of the performance gain comes from key correction ($\Delta K^{S}$).
The ``rotate'' arm preserves this gain, while ``rescale'' performs worse than disabling correction.
These results suggest that SGC improves query-slot matching primarily by adjusting slot-key directions.

\section{Discussion and Conclusion}
We introduced SD-Cache, a learnable memory consolidation framework that maintains a budget-constrained KV cache through selective and state-guided updates.
We then proposed an anchor-pooling module that compresses long KV sequences into a fixed number of memory slots while preserving key-value correspondence.
We further introduced state-guided correction to refine the pooled KV representations using global key-value associations accumulated from the original KV stream, supporting access to relevant information during query decoding.
Our method demonstrated competitive performance and computational cost compared with baselines on retrieval-augmented question answering and long-context benchmarks, covering tasks such as question answering, needle-in-a-haystack, variable tracing and aggregation.
These results suggest that SD-Cache can be a valuable step towards learnable memory consolidation for LLMs.

\textbf{Limitations and Future Work.}
% A current limitation of our method is that it does not explicitly search for or learn an optimal layer-wise memory allocation, which may further improve the framework’s performance. 
% Another promising direction is to explore more sophisticated reward functions for the consolidation system, allowing memory updates to better reflect downstream task requirements. Future work will investigate both directions.
%
A current limitation is that our method does not explicitly optimise layer-wise memory allocation, and maintaining performance across multi-step streaming updates remains challenging.
Future work will explore adaptive allocation and more sophisticated reward functions to align memory updates with downstream task requirements. We will also investigate hybrid approaches combining learned consolidation with training-free token selection, alongside more robust update strategies and training data tailored to multi-step writing.

% \clearpage

% \section{Preparing PostScript or PDF files}

% Please prepare PostScript or PDF files with paper size ``US Letter'', and
% not, for example, ``A4''. The -t
% letter option on dvips will produce US Letter files.

% Consider directly generating PDF files using \verb+pdflatex+
% (especially if you are a MiKTeX user).
% PDF figures must be substituted for EPS figures, however.

% Otherwise, please generate your PostScript and PDF files with the following commands:
% \begin{verbatim}
% dvips mypaper.dvi -t letter -Ppdf -G0 -o mypaper.ps
% ps2pdf mypaper.ps mypaper.pdf
% \end{verbatim}

% \subsection{Margins in LaTeX}

% Most of the margin problems come from figures positioned by hand using
% \verb+\special+ or other commands. We suggest using the command
% \verb+\includegraphics+
% from the graphicx package. Always specify the figure width as a multiple of
% the line width as in the example below using .eps graphics
% \begin{verbatim}
%    \usepackage[dvips]{graphicx} ...
%    \includegraphics[width=0.8\linewidth]{myfile.eps}
% \end{verbatim}
% or % Apr 2009 addition
% \begin{verbatim}
%    \usepackage[pdftex]{graphicx} ...
%    \includegraphics[width=0.8\linewidth]{myfile.pdf}
% \end{verbatim}
% for .pdf graphics.
% See section~4.4 in the graphics bundle documentation (\url{http://www.ctan.org/tex-archive/macros/latex/required/graphics/grfguide.ps})

% A number of width problems arise when LaTeX cannot properly hyphenate a
% line. Please give LaTeX hyphenation hints using the \verb+\-+ command.

\subsection*{AI use statement}
In this work, we used generative AI tools to implement and edit Python code for the proposed method, create or modify scientific figures, prototypes of methodology diagrams, visualise examples and edit the manuscript to improve its clarity and readability. We also used AI tools to retrieve and identify related works.
We did not use generative AI tools to generate synthetic datasets or to develop theoretical models. No other required-disclosure uses were applicable to this work. All AI-assisted outputs, including text, code, and figures, were reviewed by the authors, and the generated or modified code was tested for correctness. The authors take full responsibility for the final content of this work, including all text, claims, code, and other artefacts produced with the assistance of generative AI.

\subsection*{Reproducibility statement}
We support reproducibility by providing a detailed description of SD-Cache and its components in Section~\ref{sec:method}. The appendix documents the model training details, hyperparameters and additional evidence used in our experiments. We also report the backbone models, datasets, memory budgets, and evaluation metrics for each benchmark. We will publicly release the source code, configuration files, and training and evaluation scripts in the camera-ready version.



\bibliography{iclr2027_conference}
\bibliographystyle{iclr2027_conference}

\appendix

\section{Additional Details}
\subsection{Implementation Details} \label{aap:imple_details}
Our consolidation network follows the Perceiver architecture: the $M$ slot queries cross-attend to the key-value tensor through $8$ pre-norm Perceiver like blocks, each with $8$ cross-attention heads and a $4\times$ FFN expansion, operating at the backbone's key-value width. 
A single anchor pooling module (APM) is shared across all layers. Each layer also has a lightweight projector comprising copies of its frozen key and value projections, each initialised from the corresponding original weights and augmented with rank-$16$ LoRA adapters ($\alpha=32$).
This amounts to approximately $142$M trainable parameters, $1.7\%$ of an 8B backbone, which remains frozen throughout. We set fixed correction strengths $\alpha_v=\alpha_k=0.1$.
The computed memory slots are kept in pre-RoPE space; the $i$-th slot is rotated at the position of the anchor token it was initialised from, and the query and generated tokens are positioned directly after the $M$ slots.
We train for five epochs using AdamW~\citep{Loshchilov2017FixingWD} with a learning rate of $2\times10^{-4}$, a linear learning rate schedule with $3\%$ warmup, and bf16 precision. Depending on the context length, we use effective batch sizes of $128$ or $64$, with corresponding per-device batch sizes of $4$ or $1$ and gradient accumulation.
Training runs under DeepSpeed ZeRO-2 on four RTX 6000 Ada GPUs; the per-layer trunk calls are gradient-checkpointed, which leaves the computed gradients unchanged and is what fits the largest memory budget into $48$\,GB.
For the experiments, we do not include KV-Distill~\citep{chari2025kv} in our RULER~\citep{hsieh2024ruler} comparison because no public implementation was available at the time of our experiments, making faithful reproduction difficult. We also omit STILL~\citep{o2026still} because its peak training memory requirements (e.g., 8 $\times$ NVIDIA H200 GPUs) exceeded our available hardware budget.

\subsection{BD-KV Computation}
\label{app:bd-kv}

We adapt the Bj{\o}ntegaard-Delta rate metric~\citep{bjontegaard2001calculation} by using the nominal retained KV fraction as the rate variable and Match as the quality variable. While BD-Rate is typically applied in image and video compression, recent work has argued for its use to evaluate LLM compression methods~\citep{gordon2026sinelora}. For each method $m$, we construct Piecewise Cubic Hermite Interpolating Polynomial (PCHIP~\citep{fritsch1980monotone}) interpolants $g_m(q)$ through the measured points $(q_i,\log_{10} r_i)$, where $q_i$ denotes Match and $r_i$ denotes the retained KV fraction. The relative KV difference at a target quality $q$ is
\begin{equation}
    \Delta_{\mathrm{KV}}(q)
    =
    100\left(
        10^{g_{\mathrm{ours}}(q)-g_{\mathrm{base}}(q)}-1
    \right).
    \label{eq:relative-kv}
\end{equation}

We restrict evaluation to the intersection of the observed quality ranges:
\begin{align}
    q_{\min}
    &= \max\left\{
        \min_i q_i^{\mathrm{ours}},
        \min_j q_j^{\mathrm{base}}
    \right\}, \\
    q_{\max}
    &= \min\left\{
        \max_i q_i^{\mathrm{ours}},
        \max_j q_j^{\mathrm{base}}
    \right\}.
\end{align}
For $q_{\max}>q_{\min}$, we compute
\begin{align}
    \overline{\delta}
    &=
    \frac{1}{q_{\max}-q_{\min}}
    \int_{q_{\min}}^{q_{\max}}
    \left[
        g_{\mathrm{ours}}(q)-g_{\mathrm{base}}(q)
    \right]\,\mathrm{d}q, \\
    \mathrm{BD\text{-}KV}
    &=100\left(10^{\overline{\delta}}-1\right).
    \label{eq:bd-kv}
\end{align}

\begin{table}[t]
    \centering
    \small
    \setlength{\tabcolsep}{4pt}
    \renewcommand{\arraystretch}{1.15}
    \caption{
        Average accuracy (\%) across the 13 RULER-4K tasks~\citep{hsieh2024ruler} under different KV cache reduction ratios, using Qwen3-8B~\citep{yang2025qwen3} as the base model. A reduction ratio of $0\%$ denotes the full, uncompressed KV cache.
    }
    \begin{tabularx}{\linewidth}{
        l
        >{\raggedright\arraybackslash}X
        rrrrr
    }
        \toprule
        \multirow{2}{*}{\textbf{Model}}
        & \multirow{2}{*}{\textbf{Method}}
        & \multicolumn{5}{c}{\textbf{KV cache reduction}} \\
        \cmidrule(lr){3-7}
        & & 0.00\% & 90.00\% & 93.75\% & 96.88\% & 98.44\% \\
        \midrule
        \multirow{6}{*}{\makecell{Qwen3\\8B}}
        & PyramidKV~\citep{cai2024pyramidkv}
        & 95.30 & 19.50 & 15.20 & 9.35 & 2.31 \\
        & SnapKV~\citep{li2024snapkv}
        & 95.30 & 19.20 & 15.22 & 9.25 & 2.33 \\
        & TOVA~\citep{oren2024transformers}
        & 95.30 & 24.70 & 11.00 & 4.09 & 3.60 \\
        & KeyDiff~\citep{park2026keydiff}
        & 95.30 & 37.90 & 25.40 & 14.61 & 8.90 \\
        & IndexMem~\citep{yang2026indexmem}
        & 95.30 & 46.20 & - & - & - \\
        \cmidrule(l){2-7}
        \rowcolor{gray!10}
        \cellcolor{white}
        & \textbf{SD-Cache (Ours)}
        & 95.30
        & \textbf{61.55}
        & \textbf{59.80}
        & \textbf{57.32}
        & \textbf{56.39} \\
        \bottomrule
    \end{tabularx}
    \label{table:ruler_qwen}
\end{table}

\begin{table}[t]
    \centering
    \small
    \setlength{\tabcolsep}{4pt}
    \renewcommand{\arraystretch}{1.15}
    \caption{
        Average accuracy (\%) across the 13 RULER-4K tasks~\citep{hsieh2024ruler} under different KV cache reduction ratios, using Mistral-7B~\citep{Jiang2023Mistral7} as the base model. A reduction ratio of $0\%$ denotes the full, uncompressed KV cache.
    }
    \begin{tabularx}{\linewidth}{
        l
        >{\raggedright\arraybackslash}X
        rrrrr
    }
        \toprule
        \multirow{2}{*}{\textbf{Model}}
        & \multirow{2}{*}{\textbf{Method}}
        & \multicolumn{5}{c}{\textbf{KV cache reduction}} \\
        \cmidrule(lr){3-7}
        & & 0.00\% & 90.00\% & 93.75\% & 96.88\% & 98.44\% \\
        \midrule
        \multirow{6}{*}{\makecell{Mistral\\7B-v0.3}}
        & PyramidKV~\citep{cai2024pyramidkv}
        & 92.10 & 19.90 & 13.71 & 8.43 & 4.10 \\
        & SnapKV~\citep{li2024snapkv}
        & 92.10 & 19.80 & 13.75 & 8.40 & 3.59 \\
        & TOVA~\citep{oren2024transformers}
        & 92.10 & 29.20 & 14.93 & 6.67 & 3.77 \\
        & KeyDiff~\citep{park2026keydiff}
        & 92.10 & - & 45.26 & 26.61 & 14.08 \\
        & IndexMem~\citep{yang2026indexmem}
        & 92.10 & 41.50 & - & - & - \\
        \cmidrule(l){2-7}
        \rowcolor{gray!10}
        \cellcolor{white}
        & \textbf{SD-Cache (Ours)}
        & 92.10
        & \textbf{52.17}
        & \textbf{47.12}
        & \textbf{44.41}
        & \textbf{37.23} \\
        \bottomrule
    \end{tabularx}
    \label{table:ruler_mistral}
\end{table}

For SD-Cache versus attention matching, the shared interval is $[59.18,66.89]\%$ Match, yielding $\mathrm{BD\text{-}KV}=-58.70\%$. Negative values indicate a lower KV budget for SD-Cache at equal quality. The solid curve shows $\Delta_{\mathrm{KV}}(q)$, while the horizontal dashed line indicates BD-KV. This summary averages log-rate differences before converting back to a percentage; it is therefore not the arithmetic mean of the plotted percentage differences.
We use the FAU-LMS \texttt{bjontegaard}~\footnote{https://github.com/FAU-LMS/bjontegaard} implementation and do not extrapolate beyond the shared quality interval. KV cost is measured by the nominal token/slot budget, rather than total device memory; model weights and auxiliary state are excluded.

\section{Additional Results}

\begin{figure}
    \centering
    \includegraphics[width=0.4\linewidth]{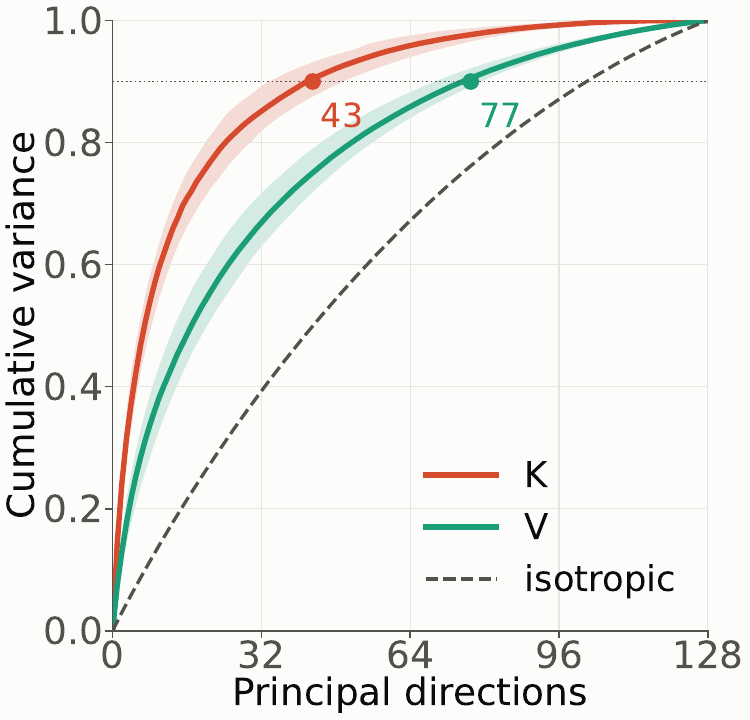}
    \caption{Visualisation of cumulative explained variance for mean-centred pre-RoPE keys (K) and values (V) in Qwen3-4B Instruct~\citep{yang2025qwen3}. Curves are averaged over 12 documents per layer-head pair, where solid lines and shaded bands show the median and inter-quartile range across 288 pairs. The dashed curve shows an isotropic Gaussian reference of the same size. Dots mark $90\%$ explained variance for K and V.}
    \label{fig:head_rank}
\end{figure}

\subsection{Low-Rank Structure within KV Heads}\label{app:apm_head_rank}

To examine the low-rank structure that motivates the anchor pooling module (APM), we analyse token variation within each KV head of Qwen3-4B Instruct.
Each document contains 768 tokens, with the initial attention-sink position removed.
We extract pre-RoPE keys from \texttt{k\_norm} and values from \texttt{v\_proj} across all 36 layers and 8 KV heads per layer ($d=128$).
For each document and head, we subtract the mean token vector from each $767 \times 128$ representation matrix and apply singular value decomposition (SVD) to the residuals.
We accumulate the squared singular values in descending order and normalise by their total sum to obtain the cumulative explained variance.
Centring is performed separately for each document because slot pooling summarises one context at a time.
For each layer-head pair, we first average the curves over the 12 documents, then report the median and inter-quartile range across the 288 pairs.
An isotropic Gaussian matrix of the same size undergoes the same centring and SVD steps as a reference.
As shown in Figure~\ref{fig:head_rank}, the median curves reach $90\%$ explained variance at $43$ directions for keys and $77$ for values, compared with $103$ for the isotropic reference.
The concentrated key variation supports the motivation for APM, while the higher effective rank of values suggests that preserving their variation may require more capacity.

\subsection{Additional Results on RULER Benchmark} \label{app:ruler_additional}
We further evaluate SD-Cache on RULER-4K~\citep{hsieh2024ruler} using two additional backbones: Qwen3-8B~\citep{yang2025qwen3} and Mistral-7B-v0.3~\citep{Jiang2023Mistral7}.
Tab.~\ref{table:ruler_qwen} and Tab.~\ref{table:ruler_mistral} report the average accuracy across the 13 RULER tasks with a 4096-token context.
SD-Cache achieves the highest reported accuracy at every non-zero cache reduction ratio on both backbones.

\textbf{On Qwen3-8B}~\citep{yang2025qwen3}, SD-Cache outperforms IndexMem~\citep{yang2026indexmem} at $90\%$ reduction and KeyDiff~\citep{park2026keydiff} at $93.75\%$ reduction, the strongest available baselines in these settings, by $15.35$~pp and $34.40$~pp, respectively.
At the two highest reduction ratios of $96.88\%$ and $98.44\%$, SD-Cache achieves accuracies of $57.32\%$ and $56.39\%$, exceeding KeyDiff by $42.71$~pp and $47.49$~pp, respectively.
Its accuracy decreases by only 3.41~pp as the reduction increases from $93.75\%$ to $98.44\%$, indicating robustness to increasingly aggressive compression.

\textbf{On Mistral-7B-v0.3}~\citep{Jiang2023Mistral7}, SD-Cache achieves accuracies of $52.17\%$ and $47.12\%$ at $90\%$ and $93.75\%$ reduction, outperforming the same respective baselines by $10.67$~pp and $1.86$~pp.
At $96.88\%$ and $98.44\%$ reduction, it achieves $44.41\%$ and $37.23\%$ accuracy, surpassing KeyDiff by $17.8$~pp and $23.15$~pp, respectively.
% Although its accuracy decreases by $9.89$~pp as the reduction increases from $93.75\%$ to $98.44\%$, compared with a 3.41~pp drop on Qwen3-8B, SD-Cache maintains the highest reported accuracy among the compression methods at all evaluated non-zero reduction ratios.
These results suggest that SD-Cache preserves useful information under severe KV cache compression across different LLM backbones.

\begin{figure}[t]
    \centering
    \includegraphics[width=1.0\linewidth]{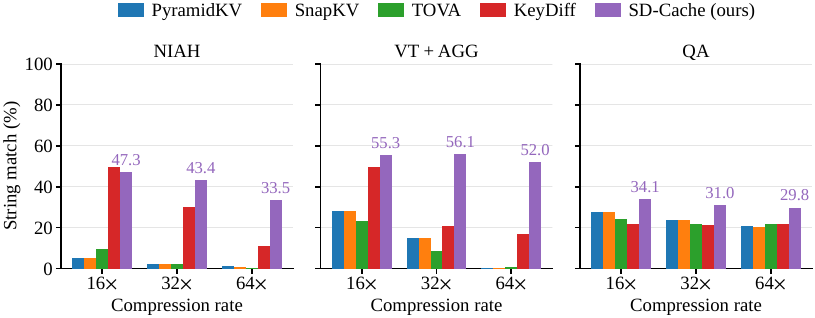}
    \vspace{-5pt}
    \caption{RULER-4K~\citep{hsieh2024ruler} performance by task group on Mistral-7B-Instruct-v0.3~\citep{Jiang2023Mistral7} under different KV cache compression rates.}
    \vspace{-10pt}
    \label{fig:ruler_taskwise_mistral}
\end{figure}
\begin{figure}[t]
    \centering
    \includegraphics[width=1.0\linewidth]{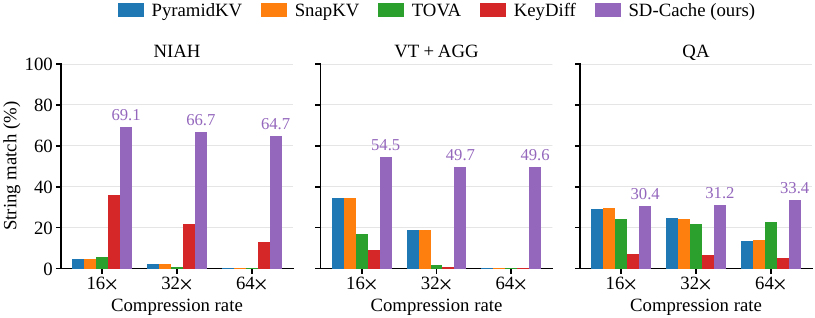}
    \vspace{-5pt}
    \caption{RULER-4K~\citep{hsieh2024ruler} performance by task group on Qwen3-8B~\citep{yang2025qwen3} under different KV cache compression rates.
    }
    \vspace{-10pt}
    \label{fig:ruler_taskwise_qwen}
\end{figure}

\textbf{Task Group Analysis.}
We group the RULER-4K tasks into four categories: needle-in-a-haystack (NIAH), question answering (QA), variable tracking (VT), and aggregation (AGG), where AGG comprises common words extraction (CWE) and frequent words extraction (FWE).
Figures~\ref{fig:ruler_taskwise_qwen} and~\ref{fig:ruler_taskwise_mistral} present results on Qwen3-8B and Mistral-7B-v0.3, respectively, with VT and AGG reported jointly as VT+AGG.
While KeyDiff shows competitive results on NIAH at $16\times$ compression, SD-Cache outperforms all evaluated baselines across all three reported groups at $32\times$ and $64\times$ compression on both backbones.
These results show that SD-Cache is generally more robust to aggressive KV cache compression than the other baselines.

\subsection{Ablation of the Novelty-Guided Write}
\label{app:novelty_guided_write}

We evaluate SD-Cache trained on RA-QA~\citep{louis2025pisco} with a Qwen3-4B-Instruct~\citep{yang2025qwen3} backbone on 948 ASQA dev questions at a $16\times$ compression ratio, using Match as the metric.
Without retraining, we test three inference-time variants: removing the novelty bias, shuffling novelty scores across tokens within each passage, and replacing the slot gate with its layer-wise mean.
To evaluate the validity of the novelty-guided write module, we additionally design a shuffling experiment that preserves the distribution of novelty scores but breaks their correspondence with the tokens, thereby testing whether assigning higher scores to the appropriate tokens matters.
The full model achieves $71.5\%$ Match, compared with $69.5\%$ without the novelty bias, $67.2\%$ with shuffled novelty scores, and $68.1\%$ with a static gate.
These results support the contributions of both token-specific novelty information and input-dependent slot gating.

\subsection{Computational-cost estimation} 
\label{app:tflops_detailed}
\begin{table}[t]
    \centering
    \small
    \setlength{\tabcolsep}{7pt}
    \renewcommand{\arraystretch}{1.08}
    \caption{
        Estimated computational cost of Llama-3.1-8B-Instruct under the RULER compression budgets.
        We use a controlled workload of 4096 context tokens, 128 question tokens and 32 output tokens, with a batch size of one and no separate system prefix.
        These costs are not averaged over RULER evaluation samples.
        $^{\dagger}$ExpectedAttention uses the AdaKV masking implementation associated with the archived accuracy results, which retains the full physical KV cache.
        Its costs are rounded analytical upper-bound estimates for successful execution paths with 1000 search iterations per call.
        $^{\ddagger}$AM-fast uses off-policy repeat queries and uniform budgets.
        Its costs include reference-query preparation and analytically estimated QR solver costs.
    }
    \vspace{-5pt}
    \begin{tabular}{llrrr}
        \toprule
        & & \multicolumn{3}{c}{Computational cost (TFLOPs) $\downarrow$} \\
        \cmidrule(lr){3-5}
        Target context-slot reduction & Method & Inference & Compact & Total \\
        \midrule
        0.00\% & Full KV & 2.74 & 65.97 & 68.71 \\
        \midrule
        \multirow{5}{*}{75.00\%} & SnapKV & 2.48 & 66.11 & 68.59 \\
         & KeyDiff & 2.48 & 65.97 & 68.45 \\
         & ExpectedAttention + AdaKV$^{\dagger}$ & 2.78 & 70.65 & 73.43 \\
         & AM-fast$^{\ddagger}$ & 2.35 & 188.75 & 191.10 \\
         & \textbf{SD-Cache (Ours)} & 2.50 & 83.46 & 85.96 \\
        \midrule
        \multirow{5}{*}{90.00\%} & SnapKV & 2.43 & 66.11 & 68.54 \\
         & KeyDiff & 2.43 & 65.97 & 68.40 \\
         & ExpectedAttention + AdaKV$^{\dagger}$ & 2.78 & 70.65 & 73.43 \\
         & AM-fast$^{\ddagger}$ & 2.30 & 172.11 & 174.41 \\
         & \textbf{SD-Cache (Ours)} & 2.45 & 79.39 & 81.84 \\
        \midrule
        \multirow{5}{*}{93.75\%} & SnapKV & 2.42 & 66.11 & 68.53 \\
         & KeyDiff & 2.42 & 65.97 & 68.39 \\
         & ExpectedAttention + AdaKV$^{\dagger}$ & 2.78 & 70.65 & 73.43 \\
         & AM-fast$^{\ddagger}$ & 2.29 & 169.90 & 172.18 \\
         & \textbf{SD-Cache (Ours)} & 2.43 & 78.27 & 80.70 \\
        \midrule
        \multirow{5}{*}{96.88\%} & SnapKV & 2.41 & 66.11 & 68.52 \\
         & KeyDiff & 2.41 & 65.97 & 68.38 \\
         & ExpectedAttention + AdaKV$^{\dagger}$ & 2.78 & 70.65 & 73.43 \\
         & AM-fast$^{\ddagger}$ & 2.27 & 168.64 & 170.91 \\
         & \textbf{SD-Cache (Ours)} & 2.42 & 77.30 & 79.73 \\
        \midrule
        \multirow{5}{*}{98.44\%} & SnapKV & 2.40 & 66.11 & 68.51 \\
         & KeyDiff & 2.40 & 65.97 & 68.37 \\
         & ExpectedAttention + AdaKV$^{\dagger}$ & 2.78 & 70.65 & 73.43 \\
         & AM-fast$^{\ddagger}$ & 2.27 & 168.21 & 170.48 \\
         & \textbf{SD-Cache (Ours)} & 2.42 & 76.81 & 79.23 \\
        \bottomrule
    \end{tabular}
    \vspace{-10pt}
    \label{tab:supp_ruler_flops}
\end{table}

In Tab.~\ref{tab:supp_ruler_flops}, we additionally report analytical estimates of computational cost under the RULER compression budgets. 
We count each multiply–accumulate as two FLOPs and treat attention products as dense rectangular matrix multiplications. 
The estimates cover major matrix operations and follow each method’s input formatting and generation procedures. They are not GPU measurements or dataset-wide averages. 
``Inference'' ($I$) denotes question pre-fill and answer generation after preparation; ``Compact'' ($C$) denotes context or representation preparation, including context pre-fill; and ``Total'' ($T$) denotes first-use cost. For a fixed context whose prepared representation can be reused unchanged across questions, the amortised cost per question over $R$ questions is $I+C/R$. This accounting does not imply that persistent reuse has been implemented for every method. 
For Llama-3.1-8B-Instruct, the KV cache of the 4096-token context occupies 0.50 GiB (128 KiB per token in bf16). Reducing the context slots by 75\%, 90\%, 93.75\%, 96.88\% and 98.44\% shrinks it to $0.125$, $0.050$, $0.031$, $0.016$ and $0.008$ GiB, respectively, i.e. 4× to 64× smaller. The only exception is ExpectedAttention + AdaKV, whose masking implementation retains the full physical cache and therefore stays at $0.50$ GiB.

\begin{figure}[t]
    \centering
    \begin{minipage}[t]{0.495\linewidth}
        \vspace{0pt}
        \centering
        \includegraphics[width=\linewidth]{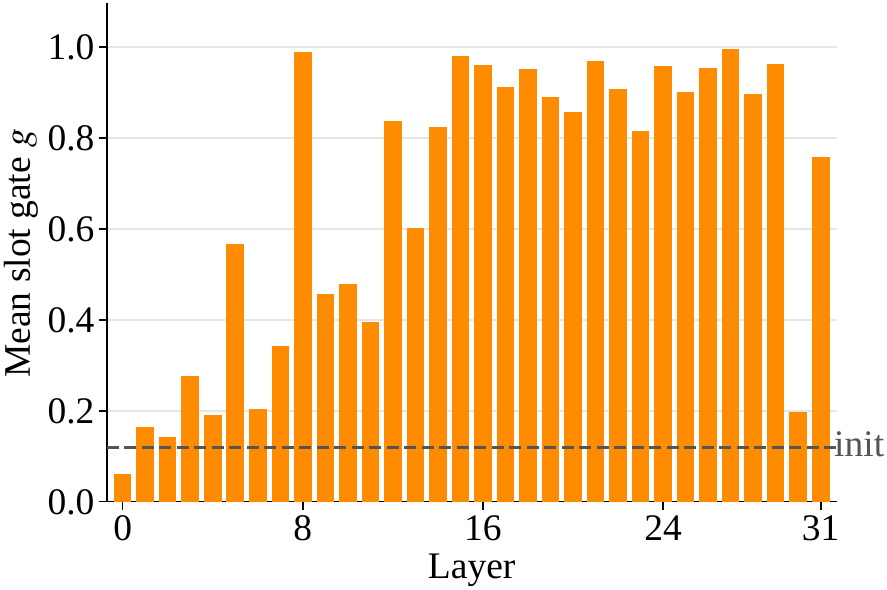}
        \caption{Distribution of estimated slot update strengths.}
        \label{fig:gate_val_dist}
    \end{minipage}
    \begin{minipage}[t]{0.495\linewidth}
        \vspace{0pt}
        \centering
        \includegraphics[width=\linewidth]{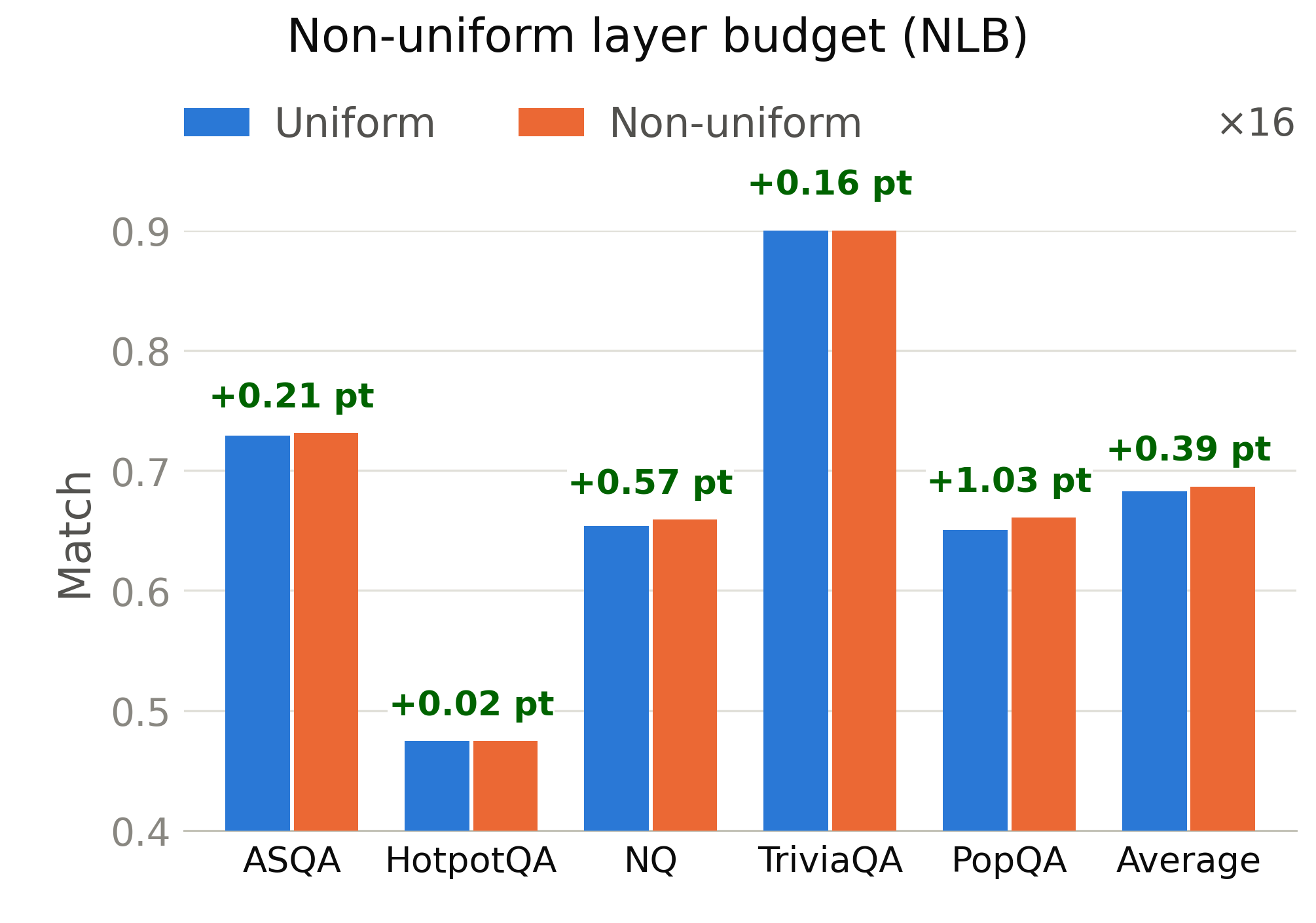}
        \caption{Uniform vs. non-uniform layer budgets.}
        \label{fig:nlb_ablation}
    \end{minipage}
    \vspace{-10pt}
\end{figure}

\subsection{Non-uniform Layer Budget} \label{app:nlb}
To visualise the contribution of the slot update strength at each layer, we examine the learned gate values $\mathbf{g}$ in Eq.~\ref{eq:apm_slot_update} across all layers. The gate magnitudes exhibit a non-uniform distribution, with several intermediate layers showing stronger updates than the early and final layers. This suggests that SD-Cache learns to vary the degree of memory update across different layers. Motivated by this layer-wise heterogeneity, we further investigate whether the memory budget itself can be distributed non-uniformly across layers.

Our SD-Cache design naturally supports non-uniform layer-wise KV cache budgets, allowing different numbers of cache slots to be assigned to different layers.
Specifically, we divide the layers into two early groups (layers 0-7 and 8-11), a middle group (layers 12-22), and a late group (layers 23-31), allocating $15.6\%$, $13.1\%$, $51.6\%$, and $19.7\%$ of the total KV cache budget to these four groups, respectively.
This allocation concentrates more capacity in the middle layers while preserving the same total budget as the uniform baseline.
As shown in Figure~\ref{fig:nlb_ablation}, non-uniform allocation yields modest but consistent improvements across all five datasets, increasing the average Match score by $0.39$~pp.
The largest gains are observed on PopQA~\citep{mallen2023not} and NQ~\citep{kwiatkowski2019natural}, with improvements of $1.03$~pp and $0.57$~pp, respectively.
These results highlight the flexibility of SD-Cache in supporting non-uniform layer-wise budgets, while leaving the exploration of more effective budget allocation strategies to future work.

\subsection{Memory Updates as the Context Changes}
\label{app:memory_update}
\begin{figure}[th]
    \centering
    \includegraphics[width=1.0\linewidth]{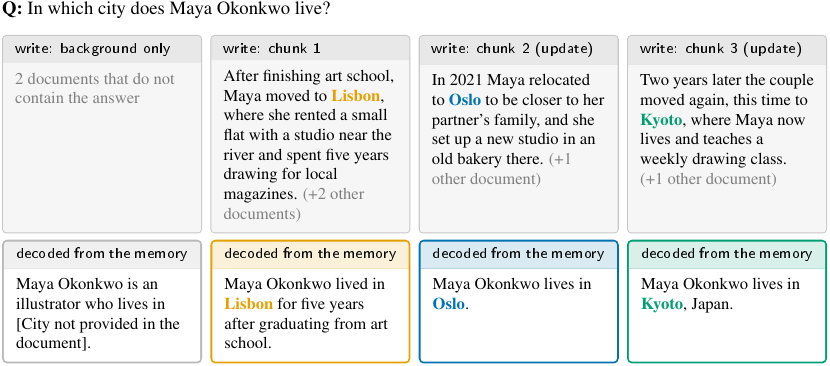}
    \vspace{5pt}
    \includegraphics[width=1.0\linewidth]{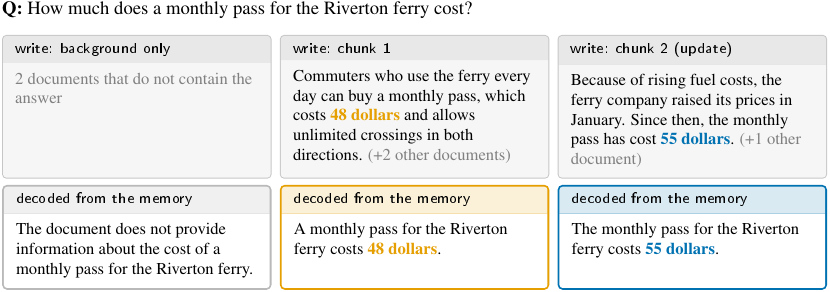}
    \vspace{5pt}
    \includegraphics[width=1.0\linewidth]{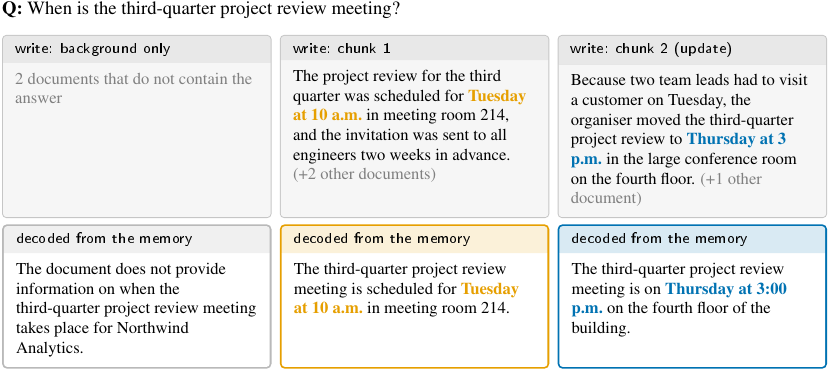}
    \caption{Visualisation of decoding from memory as the context changes. The top row shows the text written into memory at each step, with unrelated documents indicated in grey. The bottom row shows the first sentence of the answer generated by greedy decoding after each write, reproduced verbatim. Colours indicate successive states.}
      \label{fig:memory_update}
\end{figure}

We use short fictional stories to examine whether SD-Cache's answers reflect changes in the context. Each story introduces a fact and later updates it, with unrelated documents interleaved.
We first write background documents that do not contain the answer into a fixed memory of $M{=}40$ slots.
The first story chunk contains background information, the initial fact and two unrelated documents; each subsequent chunk contains one update and one unrelated document. All chunks are written into the same slots.
After each write, the frozen decoder answers the question using only the current memory.
Figure~\ref{fig:memory_update} shows the text written at each step and the corresponding decoded answer. This example illustrates SD-Cache's ability to adaptively update its memory as the context changes.

\subsection{Consolidated Memory Visualisation}
\begin{figure}[ht]
    \centering
    \includegraphics[width=1\linewidth]{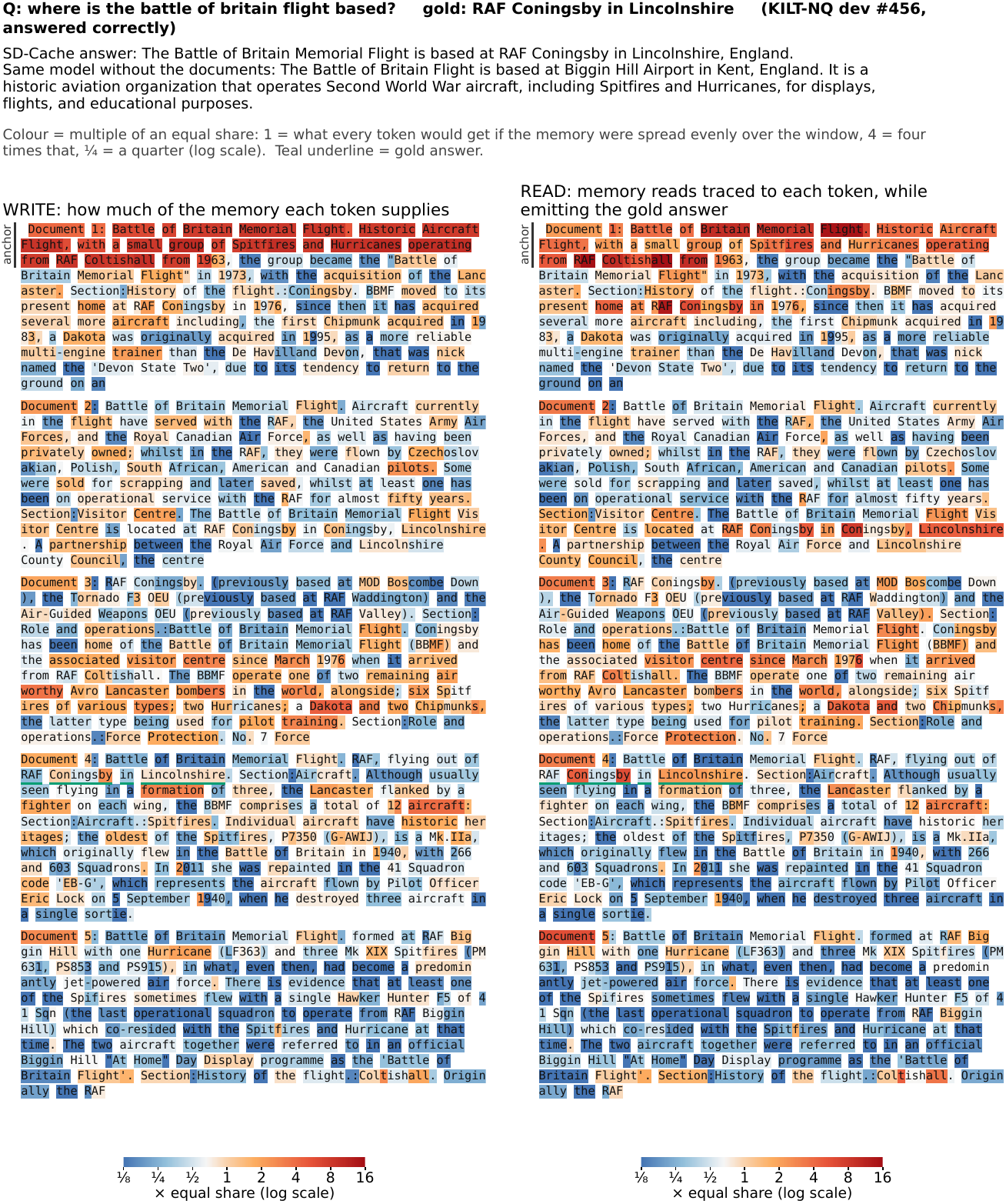}
    \caption{Visualisation of the information SD-Cache stores in memory and reads when answering question $\#456$ from the KILT-NQ dev set.}
    \label{fig:read_write_visual_456}
\end{figure}

\begin{figure}[ht]
    \centering
    \includegraphics[width=1\linewidth]{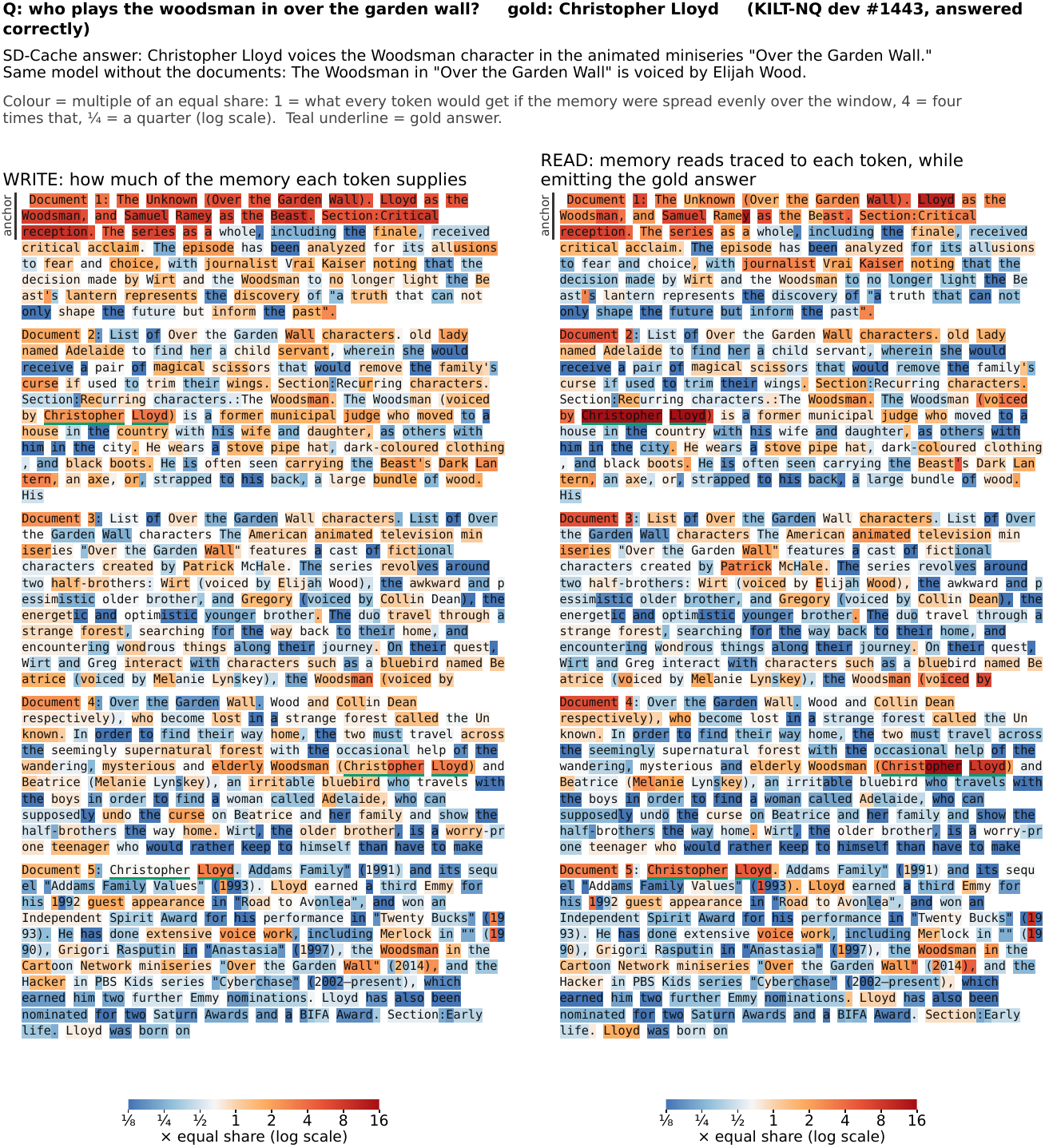}
    \caption{Visualisation of the information SD-Cache stores in memory and reads when answering question $\#1443$ from the KILT-NQ dev set.}
    \label{fig:read_write_visual_1443}
\end{figure}

To examine what SD-Cache stores and how the frozen decoder uses it, we trace memory writes and decoder reads back to the context tokens.
For \emph{WRITE}, we sum each token's contribution across slots using the anchor weights $1-g$ and pooling weights $gW$ in Eq.~(8).
For \emph{READ}, we collect the decoder's attention to the $M$ slots while generating the answer tokens, map query heads to their corresponding KV heads, and propagate the attention through the same pooling weights. 
As shown in Fig.~\ref{fig:read_write_visual_456} and Fig.~\ref{fig:read_write_visual_1443}, the READ maps place greater emphasis on answer-related content than the WRITE maps. These examples illustrate how SD-Cache consolidates information without knowing the question and allows the decoder to selectively access relevant content during decoding.

\end{document}